\documentclass[acmsmall,screen]{acmart}

\AtBeginDocument{%
  }

\setcopyright{acmlicensed}
\copyrightyear{2025}
\acmYear{2025}
\acmDOI{XXXXXXX.XXXXXXX}

\acmJournal{JACM}
\acmVolume{37}
\acmNumber{4}
\acmArticle{111}
\acmMonth{8}
\usepackage{algorithmic}
\usepackage{algorithm}
\usepackage{multirow}
\usepackage{colortbl}
\usepackage[caption=false,font=normalsize,labelfont=sf,textfont=sf]{subfig}
\usepackage{listings}
\lstdefinestyle{mystyle}{
    keywordstyle= \color{blue!70},			
    commentstyle= \color{red!50!green!50!blue!50},		
    numberstyle=\tiny\color{codegray},		
    stringstyle=\color{codepurple},
    basicstyle=\ttfamily\footnotesize,
    breakatwhitespace=false,         
    breaklines=true,		
    captionpos=b,                    
    keepspaces=true,                 
    numbers=left,		
    numbersep=5pt,                  
    showspaces=false,                
    showstringspaces=false,		
    showtabs=false,                  
    tabsize=2,
    frame=trbl,		
}

\begin{document}

\title{Uncovering Hidden Connections: Iterative {Search} and Reasoning for Video-grounded Dialog}

\author{Haoyu Zhang}
\orcid{0000-0002-3896-170X}
\affiliation{%
  \institution{School of Computer Science and Technology, Harbin Institute of Technology (Shenzhen)}
  \city{Shenzhen}
  \state{Guangdong}
  \country{China}
}
\affiliation{%
  \institution{Pengcheng Laboratory}
  \city{Shenzhen}
  \state{Guangdong}
  \country{China}
}
\email{zhang.hy.2019@gmail.com}

\author{Meng Liu}
\authornote{Both authors are corresponding authors.}
\affiliation{%
  \institution{School of Computer Science and Technology, Shandong Jianzhu University}
  \city{Jinan}
  \state{Shandong}
  \country{China}}
\email{mengliu.sdu@gmail.com}

\author{Yisen Feng}
\affiliation{%
  \institution{School of Computer Science and Technology, Harbin Institute of Technology (Shenzhen)}
  \city{Shenzhen}
  \state{Guangdong}
  \country{China}
}
\email{23S051028@stu.hit.edu.cn}

\author{Yaowei Wang}
\affiliation{%
  \institution{Pengcheng Laboratory}
  \city{Shenzhen}
  \state{Guangdong}
  \country{China}
}
\affiliation{%
  \institution{School of Computer Science and Technology, Harbin Institute of Technology (Shenzhen)}
  \city{Shenzhen}
  \state{Guangdong}
  \country{China}
}
\email{wangyw@pcl.ac.cn}

\author{Weili Guan}
\affiliation{%
  \institution{School of Computer Science and Technology, Harbin Institute of Technology (Shenzhen)}
  \city{Shenzhen}
  \state{Guangdong}
  \country{China}
}
\email{honeyguan@gmail.com}

\author{Liqiang Nie}
\authornotemark[1]
\affiliation{%
  \institution{School of Computer Science and Technology, Harbin Institute of Technology (Shenzhen)}
  \city{Shenzhen}
  \state{Guangdong}
  \country{China}
}
\email{nieliqiang@gmail.com}

\renewcommand{\shortauthors}{Zhang et al.}

\begin{abstract}
In contrast to conventional visual question answering, video-grounded dialog necessitates a profound understanding of both dialog history and video content for accurate response generation. Despite commendable progress made by existing approaches, they still face the challenges of incrementally understanding complex dialog history and assimilating video information. In response to these challenges, we present an iterative search and reasoning framework, which consists of a textual encoder, a visual encoder, and a generator. Specifically, we devise a path search and aggregation strategy in the textual encoder, mining core cues from dialog history that are pivotal to understanding the posed questions. Concurrently, our visual encoder harnesses an iterative reasoning network to extract and emphasize critical visual markers from videos, enhancing the depth of visual comprehension. Finally, we utilize the pre-trained GPT-2 model as our answer generator to decode the mined hidden clues into coherent and contextualized answers. Extensive experiments on three public datasets demonstrate the effectiveness and generalizability of our proposed framework.
\end{abstract}



\maketitle

\section{Introduction}
With the rapid proliferation of multimedia data, {research attention has focused on vision-language tasks, proving their pivotal role in academic research~\cite{zhang2021multimodal,9525823,8845771, ALOIMONOS201542, 6126320, xiang2024dkdm}}. Notable advancements have been realized in fields such as Visual Question Answering (VQA)~\cite{liang2019focal,li2021adversarial, yang2022empirical, pmlr-v235-zhang24aj,zhang2024hcqa}, Visual Dialog~\cite{9444809, chen2022utc,DBLP:journals/tois/ChenSJLHN24}, and Video Captioning~\cite{seo2022end, lin2022swinbert}. Central to these tasks is the aim to refine machine capabilities in understanding visual content and to convey these understandings linguistically in ways intelligible to humans~\cite{VFD,cheng2024diffusion,NEURIPS2024_288b63aa,pmlr-v235-lv24a}. 

\begin{figure}[t]
  \centering
  \includegraphics[width=0.65\linewidth]{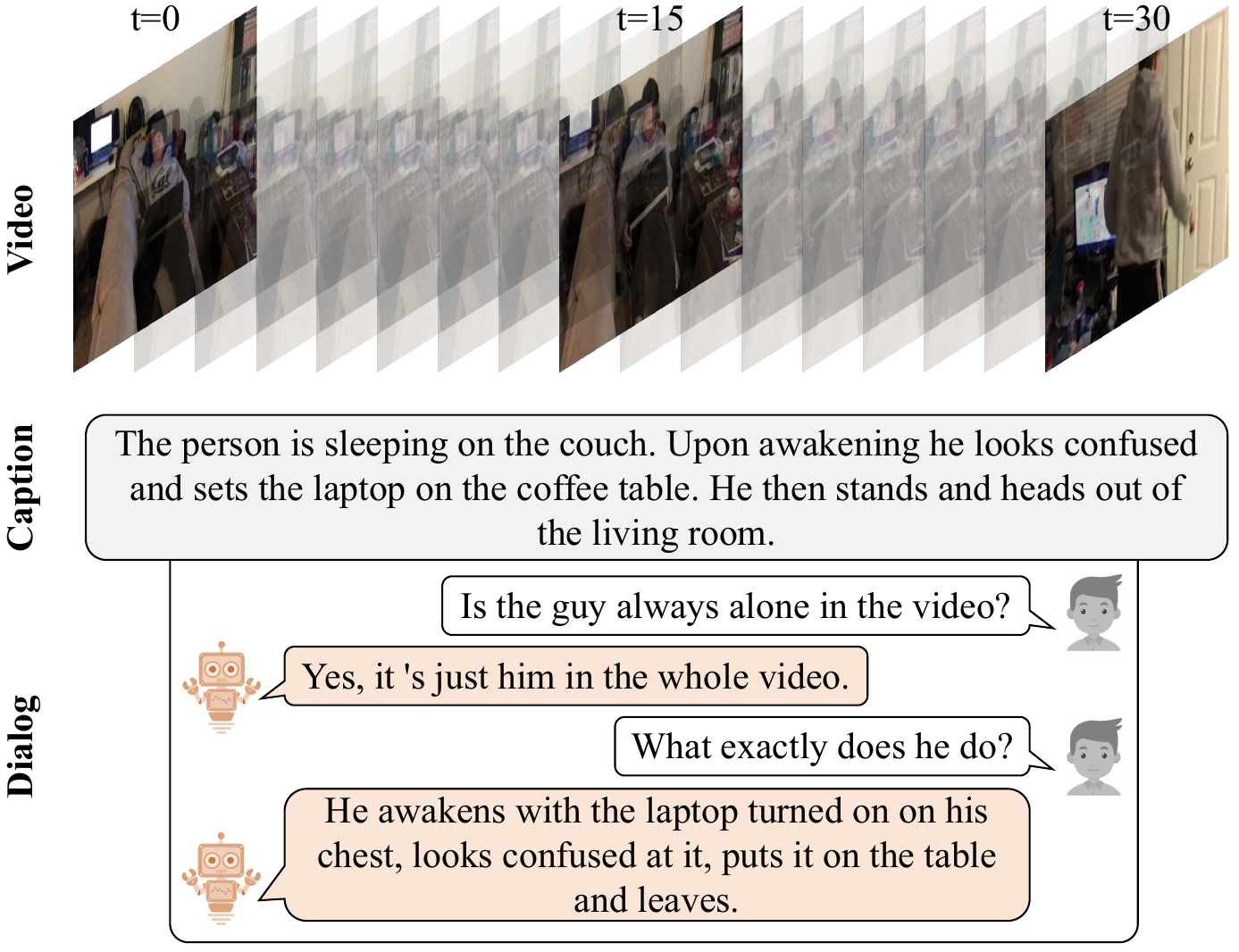}
  \caption{An illustration of video-grounded dialog. }\label{fig0}
\end{figure}

An emerging domain of significance is video-grounded dialog\footnote{\url{https://video-dialog.com/}.}, sometimes referred to as audio-visual scene-aware dialog by certain academicians~\cite{hori2019end,le2019multimodal,li2021bridging,9585294}. 
As illustrated in Fig.~\ref{fig0}, this rapidly evolving domain mandates the creation of an agent adept at conducting coherent dialogs rooted in video data, which is a monumental leap toward genuine human-computer synergy. {The implications of video-grounded dialogue extend widely, enhancing virtual personal assistants and creating innovative solutions for visually impaired individuals, highlighting its significant academic relevance~\cite{li2024optimus,li2025optimus2,li2023unisa}.}
Spurred by advancements in deep learning, the domain of video-grounded dialog is witnessing considerable traction. {Existing techniques in this domain can be methodically categorized into:}

1) \textbf{Textual Information Modeling}. {This branch emphasizes modeling textual elements, primarily involving question inquiries and their preceding contexts}~\cite{kim2021structured, le2021learning,le2021vgnmn}. For instance, Le \textit{et al.}~\cite{le2021learning} {devised a semantic graph based on the lexical components of question-answer pairs, leading to a model proficient in forecasting reasoning pathways on the graph. This method is adept at extracting question-specific data from dialog contexts.} Similarly, Kim \textit{et al.}~\cite{kim2021structured} {employed the Gumbel-Softmax technique to identify dialog content that is most pertinent to a given question, enhancing question representation in the process.}

2) \textbf{Visual Information Modeling}. {This kind of method highlights visual elements, primarily focusing on video frames and notable objects}~\cite{geng2021dynamic, le2020bist,9897613,10147255}. As a case in point, Le \textit{et al.}~\cite{le2020bist} propounded a bidirectional model that encapsulates both spatial-to-temporal relations and their converse, {aiming to mine the complex spatial and temporal cues embedded in the video.} Concurrently, Geng \textit{et al.}~\cite{geng2021dynamic} advocated representing videos via spatio-temporal scene graphs, capturing salient audio-visual cues and their semantic interrelations effectively. 

3) \textbf{Decoder Generation}. {This section emphasizes the formulation of answer derivations~\cite{le2020multimodal, le2020video,yoon-etal-2022-information,DBLP:journals/tois/FengTZZ23}. For example, Le \textit{et al.}~\cite{le2020video} extended the established functionalities of the GPT-2 architecture, discarding conventional auto-regressive decoding strategy.} Complementing this domain, the integration of pointer networks~\cite{le2020multimodal} has amplified the generative capabilities of dialog systems.

Despite the significant progress observed in the domain of video-grounded dialog, {there are still critical challenges that  demand meticulous consideration}: 
1) \textbf{Dialog History Understanding}.
Notable recent studies~\cite{le2021vgnmn, le2021learning} predominantly focus on dialog history segments that align directly with user queries. {Such a perspective can inadvertently exclude broader dialog contexts, leading to the omission of significant information.} The inherent lack of transparency in these methods further impedes a clear understanding of the decision-making process of the model. {Hence, the imperative lies in developing a superior textual encoder capable of thoroughly analyzing dialog histories while improving model interpretability.}
2) \textbf{Visual Content Understanding}.
{While several works~\cite{geng2021dynamic} endeavor to extract visual signals through complex graph structures, the importance of user-posed questions in steering visual comprehension is frequently disregarded.} This omission leads to a failure to capture question-specific insights, thereby limiting the capacity to mine meaningful visual information. { Additionally, the current trend of depending solely on a singular extraction process~\cite{9897613,10147255} may be inadequate considering the multi-faceted nature of video data.} Therefore, an approach that harmonizes user queries with iterative reasoning is crucial for a nuanced comprehension of visual content.

In response to the outlined challenges, we introduce the Iterative {Search and Reasoning (ISR)} framework for video-grounded dialog, detailed in Fig.~\ref{fig1}. This novel approach aims to address the limitations of current methods by integrating an explicit history modeling strategy alongside an iterative reasoning network. To be specific, {we implement a clear and structured path search and aggregation strategy, aimed at uncovering important information related to user queries within the dialogue context, thereby improving question comprehension. Concurrently, to harness the latent visual semantics in videos, a multimodal iterative reasoning network is devised. This architecture enables robust iterative reasoning processes, progressively refining the comprehension of visual content and consequently bolstering generation capabilities. The effectiveness of the ISR method is demonstrated through extensive experiments on three public datasets.} To advance collaborative research, the codebase is available at \textit{\url{https://github.com/Hyu-Zhang/ISR}}.


The contributions of our work can be highlighted in threefold:
{
\begin{itemize}
\item 
We present a novel approach for video-grounded dialog that synergizes the modeling of dialog history and visual content into an integrated framework. This harmonization enhances the quality and accuracy of dialog generation.
\item We design an interpretable path search and aggregation strategy, which enables the effective extraction of crucial information from the preceding dialog context, thereby enhancing the comprehensive understanding of user intention.
\item We devise a multimodal iterative reasoning network that leverages user queries. This network serves to incrementally refine the comprehension of visual content, leading to more precise answer generation.
\end{itemize}}

\section{Related Work}

\subsection{Visual Dialog}
Visual dialog represents a sophisticated extension of the VQA framework, requiring multiple rounds of sequential interactions. {The fundamental goal of this paradigm is to facilitate a seamless integration of cross-modal information derived from various inputs, culminating in the formation of a coherent textual response. Current methods within this domain may be broadly grouped into three distinct paradigms}: attention mechanism-based strategies, graph network-based techniques, and approaches reliant on pre-trained models.

\textbf{Attention Mechanism-based Approaches}. These approaches deploy attention mechanisms to pinpoint the visual regions that bear the highest relevance to the posed question. {The underlying process requires an acute understanding of the immediate context of the inquiry, as well as an insight into the historical dialog trajectory.} A salient example of this approach is the Recursive visual Attention (RvA) model, formulated by Niu \textit{et al.}~\cite{niu2019recursive}. Through iterative refinement of visual attention and retrospective analysis of dialog sequences, {this model adeptly addresses co-reference ambiguities.} Further complementing this area, Yang \textit{et al.}~\cite{yang-etal-2021-seqdialn} have innovated two Sequential Dialog Networks (SeqDialN). These networks artfully separate the complexity inherent in multimodal feature fusion from the inference process, thereby simplifying the design and implementation of the inference engine.

\textbf{Graph Network-based Techniques}. {Recognizing the pivotal role of seamless information integration, the academic community is increasingly embracing graph network techniques~\cite{9619769}.}  For instance, Schwartz \textit{et al.}~\cite{schwartz2019factor} brought forth the Factor Graph Attention (FGA) architecture. This methodology enables the meticulous construction of graphs atop utility representations, subsequently discerning their interconnections.  In a parallel vein, Guo \textit{et al.}~\cite{guo2020iterative} introduced the Context-Aware Graph (CAG) neural structure, characterized by the dynamic interdependencies exhibited by individual nodes within the graph. Uniquely, this paradigm ensures that only nodes bearing the utmost relevance are integrated, yielding a refined, context-sensitive relational graph.

\textbf{Pre-trained Model-dependent Methods}. {The contemporary era of machine learning has been marked by the rise of pre-trained models, praised for their ability to seamlessly integrate diverse informational modalities.} Anchoring on this insight, Wang \textit{et al.}~\cite{wang-etal-2020-vd} delineated a holistic strategy that harnesses pre-trained architectures to adroitly weave together visual and dialogual elements. Mirroring this trajectory, Nguyen \textit{et al.}~\cite{10.1007/978-3-030-58586-0_14} unveiled the Light-weight Transformer for Many Inputs (LTMI), a pioneering neural framework harnessing an agile transformer mechanism to masterfully orchestrate interactions across diverse utilities.

However, the nuances and complexities of video-grounded dialog task preclude the direct application of methodologies cultivated for image-based visual dialog. Video-grounded dialog manifests longer utterance sentences, additional dialog rounds, and a more intricate context that demands sophisticated comprehension~\cite{alamri2018audio}. These attributes compound the challenge, rendering the direct transposition of existing visual dialog techniques impractical and inefficacious.

\subsection{Video-grounded Dialog}
In recent years, the academic community has observed an escalating focus on video-grounded dialog systems. Noteworthy challenges, including DSTC7~\cite{alamri2018audio} and DSTC8~\cite{hori2020audio}, have been instrumental in formulating benchmarks in this domain. {This burgeoning field has subsequently witnessed the advent of numerous sophisticated methodologies to address its inherent complexities.} For example, Hori \textit{et al.}~\cite{hori2019end} pioneered an approach employing an LSTM-driven encoder-decoder framework, emphasizing multimodal attention to harmoniously integrate textual and visual cues. Expanding on this work, Le \textit{et al.}~\cite{le2019multimodal} presented the Multimodal Transformer Network (MTN), devised specifically to encode visual content while concurrently integrating diverse data modalities. Similarly, innovative contributions like the Bi-directional Spatio-Temporal reasoning model (BiST)\cite{le2020bist} have made strides in extracting both visual and spatial cues from videos, thus elucidating the interconnectedness of text and visual sequences in multidimensional aspects. 
In pursuit of refining co-referential understanding, Kim \textit{et al.}~\cite{kim2021structured} introduced the Structured Co-reference Graph Attention model (SCGA), laying the foundation for a graph-based approach deeply rooted in multimodal co-referencing techniques. In tandem with this, Geng \textit{et al.}~\cite{geng2021dynamic} advanced the Spatio-Temporal Scene Graph Representation learning technique (STSGR)—an innovative proposition capturing the semantic crux of videos through scene graphs, all the while maintaining efficient encoding.

The ascendance of pre-trained language models in myriad natural language processing domains is unmistakable. Recent efforts are directed toward fusing these models with video-grounded dialog systems. To this end, Le \textit{et al.}~\cite{le2020video} advanced the VGD-GPT framework, leveraging the capabilities of the well-regarded GPT-2 model, thus transforming video-grounded dialog generation into a more fluid sequence-to-sequence task.
Building on this progression, Li \textit{et al.}~\cite{li2021bridging} incorporated a multi-task learning approach with pre-existing language models, culminating in the inception of the Response Language Model (RLM). Recognizing the hurdles of text hallucination in text generation, Yoon \textit{et al.}~\cite{yoon-etal-2022-information} crafted the Text HallucinAtion Mitigating (THAM) architecture, integrating Text Hallucination Regularization (THR) loss into expansive pre-trained language models. In a stride towards achieving nuanced dialog generation, Zhao \textit{et al.}~\cite{zhao-etal-2022-collaborative} advocated for a Multi-Agent Reinforcement Learning (MARL) methodology, accentuating sophisticated collaborations in cross-modal analysis. With the advent of {Large Video-Language Models (LVLMs)}, such as Video-LLaMA~\cite{zhang2023video} and Otter~\cite{li2023otter}, it is evident that this domain continues to be at the forefront of research attention.

To sum up, within the domain of history modeling, a prevailing trend involves the deployment of intricate graph networks to discern semantic nuances intertwined within historical data. However, these methods often compromise on interpretability. In contrast, our approach embraces a streamlined path search and aggregation mechanism, offering superior interpretability. Additionally, current methodologies largely overlook the iterative interplay between videos and questions. We address this oversight, by devising an innovative iterative reasoning network that prioritizes this symbiotic relationship.

\begin{figure*}[t]
  \centering
  \includegraphics[width=0.99\linewidth]{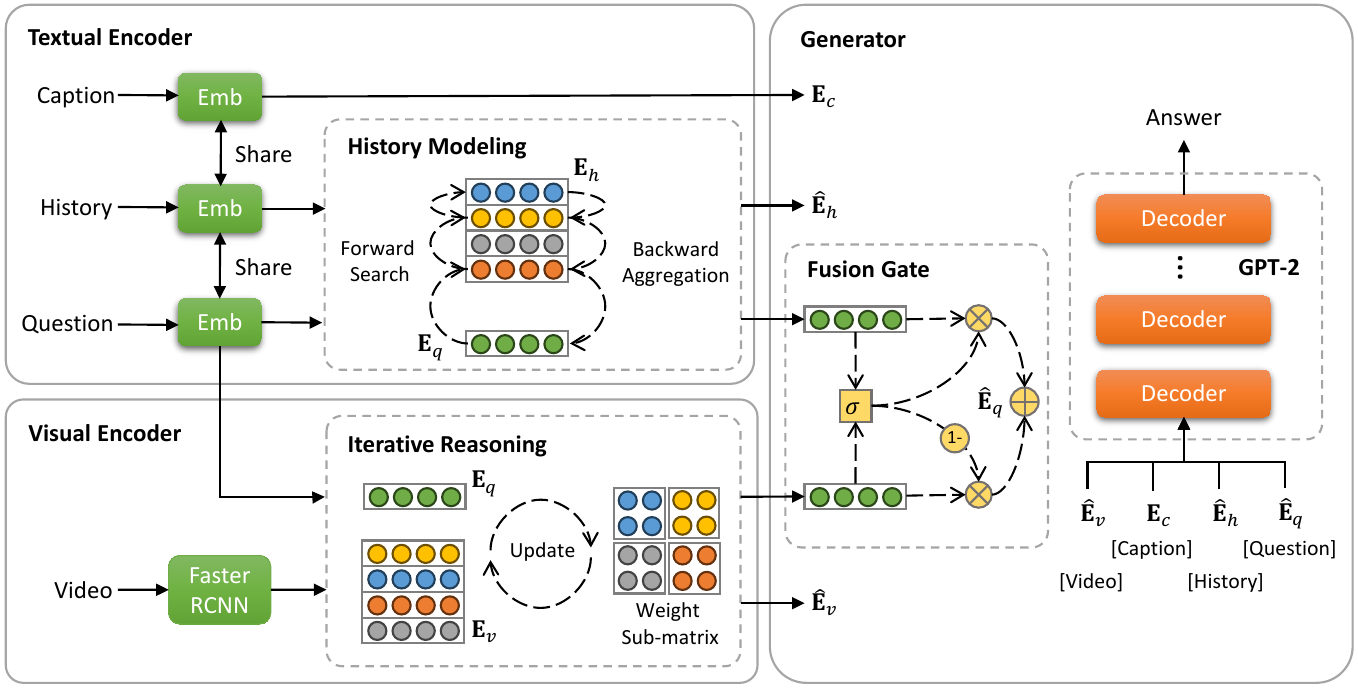}
  \caption{Illustration of our proposed framework. It consists of the textual encoder, visual encoder, and generator.}\label{fig1}
\end{figure*}

\section{Methodology}
As illustrated in Fig.~\ref{fig1}, our model consists of three components: a textual encoder, a visual encoder, and a generator. The textual encoder is designed to perform interactions between questions and history contexts and output enhanced representations. Meanwhile, the visual encoder is utilized to implement progressive reasoning between questions and videos, enhancing video content comprehension. Finally, the generator combines multiple input representations to produce system answers. In the following sections, we will elaborate on each component in detail.

\subsection{Problem Formulation}
{The objective of video-grounded dialog centers on generating a relevant answer $A_t$ through a thorough analysis of a video $V$, a caption $C$ crafted by a human to describe the video, its associated dialog history $H$, and current question $Q_t$. 
The dialog history $H$ comprises a sequence of prior questions and their answers, denoted as $(Q_1, A_1, Q_2, A_2,..., Q_{t-1}, A_{t-1})$.} 


\subsection{Textual Encoder}
\subsubsection{Preprocessing}
{The word embeddings for the given question $Q_t$, the caption $C$, and the dialog history $H$ are obtained through the same process. For clarify, we consider the question $Q_t=(w_1, w_2, ..., w_{n_q})$ as an example.} Here, an operation $f$ is applied to the vocabulary obtained from the word-to-id mapping Vocab($\cdot$). This process is represented by the following expression: 
\begin{equation}
    \mathbf{E}_q = f(\mathrm{Vocab}(w_1, w_2, ..., w_{n_q})),
\end{equation}
{where $\mathbf{E}_q \in \mathbb{R}^{{n_q}\times{d}}$ is the resulting word embeddings of question $Q_t$. The $f$ corresponds to retrieving the embedding vector associated with the id from the pre-established embedding matrix. The same step is applied to get the feature of caption $C$ and dialog history $H$, symbolized as $\mathbf{E}_c \in \mathbb{R}^{n_c \times d}$ and $\mathbf{E}_h \in \mathbb{R}^{(t-1)\times n_h \times d}$, where $n_c$ and $n_h$ respectively denote the number of words present in the caption and each utterance\footnote{{A question-answer pair is considered a utterance.}}.}

To improve the interpretability of questions drawn from the dialog history\footnote{Given that the caption inherently encapsulates the essential information, it is utilized directly in the decoding generation process without necessitating further operations.}, we introduce two preprocessing steps: co-reference resolution and dependency parsing.
Initially, the dialog history is integrated with the present question to create a unified sequence. We then utilize the AllenNLP\footnote{\url{https://allenai.org/allennlp}.} toolkit for co-reference resolution, aiming to dispel potential confusion related to pronouns. Following this, the Stanford CoreNLP package\footnote{\url{https://stanfordnlp.github.io/CoreNLP/}.} is employed to produce the dependency parse tree. {Our attention is concentrated on the parsing of dialog responses due to the inherent similarities between questions and their corresponding answers. The dependency tree delineates the syntactic relationships between sentence constituents. By adopting the subject-verb-object paradigm, we extract pertinent triplets in the 
$\langle {subject, relation, object} \rangle$
format.} This conversion facilitates the representation of both the question and dialog history as sets of triplet pairs, denoted as 
$\mathbf{E}_{tri} \in \mathbb{R}^{t\times 3 \times d}$.

\begin{figure}[t]
  \centering
  \includegraphics[width=0.6\linewidth]{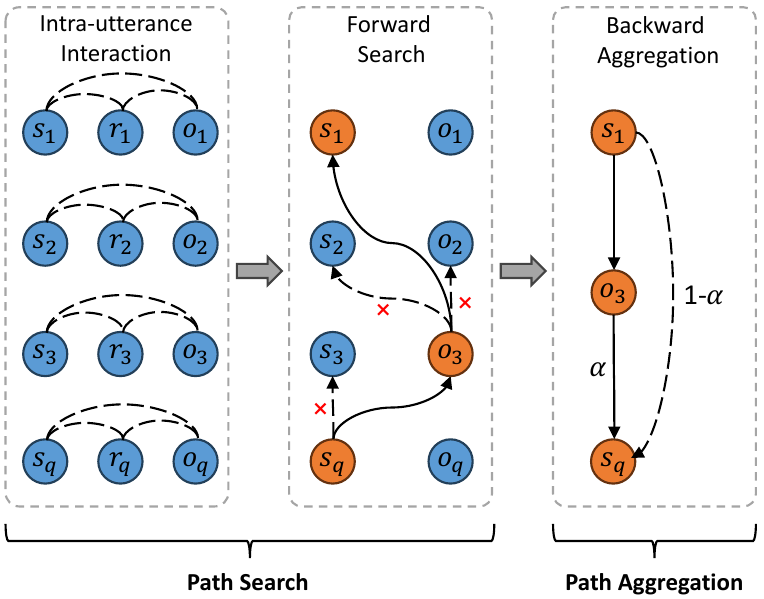}
  \caption{{Illustration of our path search and aggregation strategy.}}\label{fig:textual encoder}
\end{figure}


{Utilizing the triplet pairs $\mathbf{E}_{tri}$, we introduce a interpretable path search and aggregation strategy.} This strategy is graphically showcased in Fig.~\ref{fig:textual encoder} and methodologically detailed in Algorithm 1. {It is designed to extract pivotal semantic information from the dialog history, enriching subsequently the representation of the question, which involves two consecutive phases: path search and path aggregation.}



\subsubsection{{Path Search}}
{The objective of this phase is to construct two discrete entity paths, namely the subject path and object path.} These carefully constructed paths encompass entities derived from the dialog history, and these entities bear a substantial resemblance to the corresponding subject and object entities present in the question. {This entire process is achieved through two principal sub-processes: intra-utterance interaction and forward search.}



\textbf{(1) Intra-utterance Interaction}.
{To achieve a robust feature representation, it is imperative to incorporate the co-occurrence information of words, serving as a facilitator for the subsequent forward search step.
The refinement for the subject entity in the question is governed by the equation:}
\begin{equation}
    \overline{\mathbf{E}}_{tri}[t][0] = \mathbf{E}_{tri}[t][0] + \mathrm{AvgPool}(\mathbf{E}_{tri}[t][0:2]),
\end{equation}
where $\mathbf{E}_{tri}[t][0]$, $\mathbf{E}_{tri}[t][1]$, and $\mathbf{E}_{tri}[t][2]$ represent the feature of the subject, relation, and object entities in the question, respectively. {The AvgPool($\cdot$) denotes the average pooling operation performed on the first axis. The term $\overline{\mathbf{E}}_{tri}[t][0]$ refers to the feature of the subject entity embedded with backbone information in the question.} 
An analogous process yields the enhanced feature for the object entity, denoted as $\overline{\mathbf{E}}_{tri}[t][2]$. {By implementing the above steps on the all triplet pairs $\mathbf{E}_{tri} \in \mathbb{R}^{t\times 3 \times d}$, we derive the refined triplet representation, expressed as $\overline{\mathbf{E}}_{tri} \in \mathbb{R}^{t\times 3 \times d}$.}


\textbf{(2) {Forward Search}}.
{In this section, our focus shifts towards the execution of the forward search sub-process, which is designed to construct both the subject path and object path. 
Using the subject path as an example,
we illustrate the construction process.
For the subject entity in the question (\textit{i.e.}, the $t$-th utterance), we measure its similarity with the subject and object entities within the $(t-1)$-th utterance. The decision criterion is designed by the following mathematical formulation:}
\begin{equation}
    \begin{cases}
       S[t-1,j/2] = 1 , \; \mathrm{if} \; \epsilon_{j} > p,\\
       S[t-1,j/2] = 0 , \; \mathrm{otherwise},
    \end{cases}
\end{equation}
where $\epsilon_{j}$ represents the cosine similarity bridging the subject entity in the question and the $j$-th entity in the $(t-1)$-th utterance. {The symbol $p$ denotes a predefined selection threshold, while $S \in \mathbb{R}^{(t-1)\times 2}$ is a connection matrix.} 
Note that if both entities in the $(t-1)$-th utterance are larger than $p$, we choose the larger one, as shown in Algorithm 1.
The criterion $S[t-1,j/2] = 1$ indicates that the $j$-th entity in the $(t-1)$-th utterance resonates with the subject entity in the question. Hence, the aforementioned $j$-th entity is appointed as a pivotal reference point for the ensuing forward comparison.

{Furthermore, it is crucial to evolve from the rudimentary comparison feature, namely the feature of the subject in question $\overline{\mathbf{E}}_{tri}[t][0]$.} We propose a more advanced comparison feature, denoted by $\mathbf{h}$, which is structured as:
\begin{equation}
    \mathbf{h} = f_t(\overline{\mathbf{E}}_{tri}[t-1][j]||\overline{\mathbf{E}}_{tri}[t][0]),
\end{equation}
where $\overline{\mathbf{E}}_{tri}[t][0]$ and $\overline{\mathbf{E}}_{tri}[t-1][j]$ are the feature of the subject entity in the question and the $j$-th entity (subject or object entity) in the $(t-1)$-th utterance that correlates with the subject entity in the question, respectively. 
The concatenation operation is symbolized by $||$, while $f_t$ denotes a fully connected layer, and $\mathbf{h}$ represents the freshly synthesized feature for the ensuing round of comparison.

By perpetuating the forward comparison until reaching the inception of the dialog, a subject path rooted in the subject entity in the question is formulated. {A parallel procedure also leads to the creation of the object path.}

\begin{algorithm}[t]
\caption{{The path search and
aggregation strategy}}
\begin{algorithmic}[1]
\REQUIRE The representation of triplet pairs $\mathbf{E}_{tri}$, a fully connected layer $f_t$
\ENSURE The context-enhanced question representation $\mathbf{E}_{q}^{txt}$ and dialog history representation $\hat{\mathbf{E}}_{h}$
\STATE Establish effective representation $\overline{\mathbf{E}}_{tri}$ by Eqn. (2)
\STATE {Initialize connection matrix $S$, comparison feature $\mathbf{h}$}  
\FOR{$i \leftarrow \{0, 2\}$}
\STATE comparison point $(t,i)$, $\mathbf{h}=\overline{\mathbf{E}}_{tri}[t][i]$ 
\WHILE{$\tau \leftarrow (t-1)$ to 1}
    \STATE $\epsilon_{0}$ = Cos($\mathbf{h}, \overline{\mathbf{E}}_{tri}[\tau][0]$)
    \STATE $\epsilon_{2}$ = Cos($\mathbf{h}, \overline{\mathbf{E}}_{tri}[\tau][2]$)
    \IF{$\epsilon_{0} > p$ or $\epsilon_{2} > p$}
        \STATE $j =$ argmax($\epsilon_{0},\epsilon_{2}$)
        \STATE comparison point $(\tau,j)$
        \STATE $S[\tau,j/2]=1$, $\mathbf{h}=f_t(\overline{\mathbf{E}}_{tri}[\tau][j] || \mathbf{h})$
    \ENDIF
\ENDWHILE
\ENDFOR
\FOR{$i \leftarrow 1$ to $t$}
\STATE Obtain $\mathbf{E}_{tri}^{txt}[i][0]$ and $\mathbf{E}_{tri}^{txt}[i][2]$ based on matrix $S$ by Eqn. (5)
\ENDFOR
\STATE Output question representation $\mathbf{E}_{q}^{txt}$ and dialog history representation $\hat{\mathbf{E}}_{h}$
\end{algorithmic}
\end{algorithm}

\subsubsection{Path Aggregation}
In this section, we elaborate on the aggregation mechanism designed to use the extracted entity paths. The overarching goal is to synthesize a more refined textual representation that integrates the crux of the dialog. To achieve this, we first reverse the constructed path during Path Search. For example, in Fig.~\ref{fig:textual encoder}, if the original path is $s_q \rightarrow o_3 \rightarrow s_1$, after reversal, it becomes  $s_1 \rightarrow o_3 \rightarrow s_q$. Then to enhance long-distance modeling capability for longer paths and enrich feature representation, we add directed edges between each node and its subsequent nodes if they do not exist. For example, in the above case, a directed edge ($s_1 \rightarrow s_q$) is added from $s_1$ to its subsequent node $s_q$.

{Our strategy leverages an attention mechanism~\cite{10239469,DBLP:journals/tois/LiLN22,liu2018attentive}, ensuring that during the aggregation process, every node accumulates information from its preceding nodes based on feature similarity and distance measurement, thus refining its representation. For a clearer understanding, let's consider the case of refining the $i$-th ($i \in \{0, 2\}$) entity in the question. Its updated representation is computed as follows:
\begin{equation}
\begin{cases}
    \mathbf{E}_{tri}^{txt}[t][i] = \overline{\mathbf{E}}_{tri}[t][i] + \sum_{(k, j)\in\mathcal{N}_i}\alpha_{i,j}^{t,k}\mathbf{W}_{1}\mathbf{E}_{tri}^{txt}[k][j],\\
    \alpha_{i,j}^{t,k} = \mathrm{Softmax}(\gamma_{i,j}^{t,k}+1/(t-k)),\\
    \gamma_{i,j}^{t,k} = \overline{\mathbf{E}}_{tri}^{T}[t][i]\mathbf{W}_{1}\mathbf{E}_{tri}^{txt}[k][j],
\end{cases}
\end{equation}
where $\overline{\mathbf{E}}_{tri}[t][i]$ represents the feature of the $i$-th entity in the $t$-th utterance (the current question), {while $\mathcal{N}_i$  embodies the set of nodes connected to the $i$-th entity, \textit{i.e.}, all nodes on the same path.} Moreover, $\mathbf{E}_{tri}^{txt}[k][j]$ signifies the updated feature with aggregated antecedent information of the $j$-th neighboring entity in the $k$-th utterance. The matrix $\mathbf{W}_{1}$ functions as a parameter matrix used for transformation and $\mathbf{E}_{tri}^{txt}[t][i]$ connotes the updated feature of the $i$-th entity. An essential aspect of our strategy involves incorporating the distance between utterances into the computation of attention weights. Iteratively applying this mechanism across utterances (from $1$ to $t$) ensures that each entity feature is enhanced. It is worth noting that for nodes with zero in-degree, \textit{e.g.}, entities in the first utterance, we retain their original features.

{Following the above process, we derive a richer question representation denoted as $\mathbf{E}_{q}^{txt} \in \mathbb{R}^{n_q \times d}$.} It is worth noting that $\mathbf{E}_{q}^{txt}$ is constructed by substituting the original word features $\mathbf{E}_{q}$ with the newly refined features from the triplet pairs while leaving word features from other positions intact. Moreover, the word features in the triplet pairs, represented as $\mathbf{E}_{tri}^{txt}[:t-1]$, are regarded as the renewed dialog history representations, symbolized as $\hat{\mathbf{E}}_{h}$. {This approach not only shortens the input sequence length but also significantly minimizes disruptions caused by irrelevant words.}

\begin{figure}[t]
  \centering
  \includegraphics[width=0.7\linewidth]{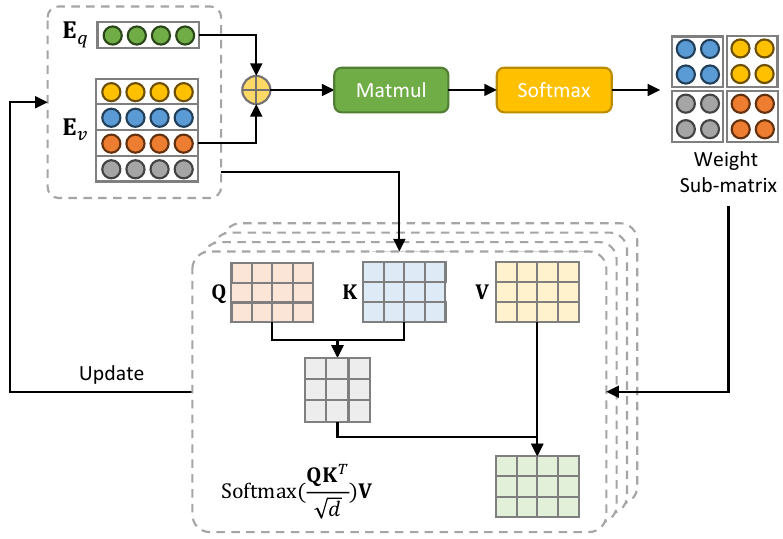}
  \caption{{Illustration of our multimodal iterative reasoning network.}}\label{fig:visual encoder}
\end{figure}

\subsection{Visual Encoder}
\subsubsection{Preprocessing}
{
Following~\cite{le2021vgnmn, kim2021structured}, we adopt Faster R-CNN model\footnote{\url{https://github.com/peteanderson80/bottom-up-attention}.} pretrained on the Visual Genome dataset to extract object features of each frame in the video, which is a renowned convolutional network model optimized for object detection task~\cite{ren2015faster,9354502,8967004}. 
Given the richness of videos in terms of the number of frames ($F$) and the plethora of objects ($O$) within these frames, we incorporate them into a single dimension, denoted as $\mathbf{E}^{'}_v\in\mathbb{R}^{{n_v}\times{2048}}$, where $n_v = F \times O$.}


{Although extraction is a crucial step, aligning these extracted features with the textual embeddings is also paramount.} Features from disparate sources often reside in different dimensional spaces; thus, a mapping mechanism becomes vital to bring congruity between them. {For this purpose, we integrate a linear layer, further enhanced with the ReLU activation function and layer normalization.} This can be mathematically expressed as:
\begin{equation}
\mathbf{E}_v = \mathrm{Norm}(\phi(f_v(\mathbf{E}^{'}_v))),
\end{equation}
{where $f_v$ acts as a fully connected layer.} The ReLU activation function, symbolized by $\phi$, introduces non-linearity, ensuring the model can learn more complex patterns. Layer normalization, denoted by $\mathrm{Norm}(\cdot)$, scales and shifts the features to ensure they have a mean of zero and a standard deviation of one. {Following this alignment procedure, the resulting feature is denoted as $\mathbf{E}_v \in \mathbb{R}^{n_v \times d}$.}

\subsubsection{Iterative Reasoning}
For a comprehensive extraction of data related to the question 
$Q_t$ from video $V$, we introduce a novel multimodal iterative reasoning network, as shown in Fig.~\ref{fig:visual encoder}. This design prioritizes cyclical modality interactions, incrementally refining the information acquired. Both modalities converge into a weight matrix, described as:
\begin{equation}
\begin{cases}
    \mathbf{E} = \mathbf{E}_q || \mathbf{E}_v,\\
    \mathbf{W} = \mathrm{Softmax}(\mathbf{E}\mathbf{E}^{T}),
\end{cases}
\end{equation}
{where $\mathbf{W}\in\mathbb{R}^{{(n_q+n_v)}\times{(n_q+n_v)}}$ represents the weight matrix, $||$ indicates concatenation operation, and Softmax($\cdot$) is performed on the last axis.} 
The matrix is divided into four specific sub-matrices, reflecting different weight components: 
$\mathbf{W}[:n_q,:n_q]$ for question-question weight, 
$\mathbf{W}[:n_q,n_q:]$ for question-video weight, 
$\mathbf{W}[n_q:,:n_q]$  for video-question weight, and 
$\mathbf{W}[n_q:,n_q:]$ for video-video weight.

{We utilize four separate attention networks~\cite{DBLP:journals/tois/ZhangCZWZLL23} without sharing parameters, each capturing distinct semantic information and enhancing the representation across different dimensions. This setup serves two advantages: 1) Each attention network can focus on learning distinct representations for different modalities, preventing information loss due to excessive feature blending. 2) A single attention mechanism may struggle to process diverse modalities efficiently, leading to information bottlenecks. Independent networks ensure that each modality gets sufficient attention without interference.}
{For example, the video-augmented question representations are derived as:}
\begin{equation}
    \begin{cases}
     \mathbf{E}_{q}^{v}= \mathbf{W}^O||_{j=1}^{J}\mathrm{head}_j,\\
    \mathrm{head}_j = \mathrm{Attention}(\mathbf{E}_{q}\mathbf{W}_{j}^{Q},\mathbf{E}_{v}\mathbf{W}_{j}^{K},\mathbf{E}_{v}\mathbf{W}_{j}^{V}),\\
    \mathrm{Attention}(\mathbf{Q},\mathbf{K},\mathbf{V})=\mathrm{Softmax}(\frac{\mathbf{Q}\mathbf{K}^T}{\sqrt{d}})\mathbf{V},
    \end{cases}
\end{equation}
{where the matrices $\mathbf{W}^O$, $\mathbf{W}_{j}^{Q}$, $\mathbf{W}_{j}^{K}$, and $\mathbf{W}_{j}^{V}$ are trainable parameters and the symbol $d$ denotes feature dimension.} The number of attention heads is depicted by $J$, while the video-enhanced question representations are illustrated by $\mathbf{E}_{q}^{v}$. Similarly, the self-enhanced question representations $\mathbf{E}_{q}^{q}$, question-enhanced video representations $\mathbf{E}_{v}^{q}$, and self-enhanced video representations $\mathbf{E}_{v}^{v}$ are obtained through the same process.

{Subsequently, we integrate intra- and inter-modality information to obtain more comprehensive representations}, as described by:
\begin{equation}
    \begin{cases}
        \mathbf{E}_{q}^{vis} = \mathrm{Sum}(\mathbf{W}[:n_q, :n_q])*\mathbf{E}_{q}^{q} +  \mathrm{Sum}(\mathbf{W}[:n_q, n_q:])*\mathbf{E}_{q}^{v},\\
        \hat{\mathbf{E}}_{v} = \mathrm{Sum}(\mathbf{W}[n_q:, :n_q])*\mathbf{E}_{v}^{q} +  \mathrm{Sum}(\mathbf{W}[n_q:, n_q:])*\mathbf{E}_{v}^{v},
    \end{cases}
\end{equation}
where $\mathrm{Sum}(\cdot)$ implies the summation function, and $\mathbf{E}_{q}^{vis}\in \mathbb{R}^{n_q \times d}$ and $\hat{\mathbf{E}}_{v}\in \mathbb{R}^{n_v \times d}$ represent the fused question and video representations, respectively. 

We recognize that a singular interaction is inadequate to fully elucidate the complex semantic signals within videos. {Hence, the described procedure can be cycled, $I$ times, iteratively refining the features.} Specifically, $\mathbf{E}_{q}^{vis}$ and $\hat{\mathbf{E}}_{v}$ are considered as new inputs in Eqn. (7) in the next iteration. {Multiple iterations appear to foster more effective representations for both the question and the video content, thus enhancing the overall system performance.} A detailed examination of this iterative process can be found in the Parameter Analysis section.



\subsection{Generator}
To deepen the understanding of inquiries, we introduce a neural architecture equipped with gating mechanisms. This architecture adeptly merges relevant information from both dialog history and visual contexts. {Specifically, we leverage a dynamic gate regulated by the initial question embeddings $\mathbf{E}_{q}$, text-rich question representations $\mathbf{E}_{q}^{txt}$, and vision-infused question representations $\mathbf{E}_{q}^{vis}$. The presentation is given as follows:}
\begin{equation}
    \mathbf{g} = \sigma(\mathbf{W}_g[\mathbf{E}_{q}||\mathbf{E}_{q}^{txt}||\mathbf{E}_{q}^{vis}]),
\end{equation}
where $\mathbf{W}_g \in \mathbb{R}^{{d}\times{3d}}$ signifies the learnable parameter matrix, the symbol $||$ represents concatenation, and $\sigma$ is the Sigmoid function. {The controlling vector $\mathbf{g} \in \mathbb{R}^{n_q \times d}$, comprising elements within the range of 0 and 1, dictates the inclusion of the necessary information, used as follows:}
\begin{equation}
    \hat{\mathbf{E}}_q = \mathbf{g} \odot \mathbf{E}_{q}^{txt} + (\mathbf{1}-\mathbf{g}) \odot \mathbf{E}_{q}^{vis},
\end{equation}
where $\odot$ is indicative of the element-wise product operation and $\hat{\mathbf{E}}_q$ denotes the ultimate question representations. This gating mechanism flexibly adjusts the influence of each modality during reasoning, ensuring a balanced and holistic query representation.

Building upon the methodology proposed by~\cite{le2021learning}, we utilize the pre-trained GPT-2 model, recognized for its advanced text generation capabilities, as the backbone for our response generator. Specifically, the generator creates responses based on the representations of the video $\hat{\mathbf{E}}_{v}$, the caption $\mathbf{E}_c$, the dialog history $\hat{\mathbf{E}}_{h}$, and the posed question $\hat{\mathbf{E}}_q$. {The combined input for the generator is structured as $(\hat{\mathbf{E}}_{v}, [\mathrm{SEP}], \mathbf{E}_c, [\mathrm{SEP}], \hat{\mathbf{E}}_{h}, [\mathrm{SEP}], \hat{\mathbf{E}}_q)$. Utilizing several transformer layers, the probability of generating a response $A_t=(w_{1}^{a}, w_{2}^{a}, ..., w_{n_a}^{a})$ is derived as:}

\begin{equation}
{
\begin{split}
    p(A_t|V, C, H, Q_t; \theta) & \approx p(A_t|\hat{\mathbf{E}}_{v}, \mathbf{E}_c, \hat{\mathbf{E}}_{h}, \hat{\mathbf{E}}_q; \theta)\\
    & =\prod_{i=1}^{n_a} p(w_{i}^{a}|w_{<i}^{a},\hat{\mathbf{E}}_{v}, \mathbf{E}_c, \hat{\mathbf{E}}_{h}, \hat{\mathbf{E}}_q;\theta),
\end{split}}
\end{equation}
where $n_a$ denotes the word number and $\theta$ signifies the parameter of the response generator.

\subsection{Optimization}
To refine our optimization strategy, we focus on reducing the negative log-likelihood associated with the actual answer. {Therefore, we employ the Maximum Likelihood Estimation (MLE)~\cite{li2021bridging,DBLP:journals/tois/LanMWGH24} loss and adopt the gradient descent method for optimization.} The expression for the loss is:

\begin{equation}
{
    \mathcal{L}_{mle} = -\sum_{i=1}^{n_a} \mathrm{log}\,    p(w_{i}^{a}|w_{<i}^{a},\hat{\mathbf{E}}_{v}, \mathbf{E}_c, \hat{\mathbf{E}}_{h}, \hat{\mathbf{E}}_q;\theta).
}
\end{equation}

\begin{table}[t]
\renewcommand{\arraystretch}{1.2}
\caption{{Details of the benchmark datasets AVSD@DSTC7, AVSD@DSTC8, and VSTAR.}}
\centering
\begin{tabular}{ccccc}
\hline
\hline
Dataset & Type &Dialogs & Turns & Words \\
\hline
AVSD@DSTC7 &Recorded Video& 11K & 202K & 1.8M\\
AVSD@DSTC8 &Recorded Video& 11K & 208K & 1.9M\\
VSTAR &TV Series& 185K & 4.6M &30.8M\\
\hline 
\hline
\end{tabular}
\label{tab:dataset}
\end{table}

\begin{table*}[t]
    \caption{Performance comparison with the latest methods on the AVSD@DSTC7 dataset. The best results are highlighted in bold and the second is underlined.}
    \renewcommand{\arraystretch}{1.1}
    \centering
    \begin{tabular}{ccccccccc}
        \hline\hline
        Method  & BLEU-1 & BLEU-2 & BLEU-3 & BLEU-4 & METEOR & ROUGE-L & CIDEr\\
         \hline
         Baseline~\cite{hori2019end} &0.626 & 0.485 & 0.383 & 0.309 & 0.215 & 0.487 & 0.746 \\
         HMA~\cite{le2019end}& 0.633 & 0.490 & 0.386 & 0.310 & 0.242 & 0.515 & 0.856 \\
         RMFF~\cite{yeh2019reactive} &0.636 & 0.510 & 0.417 & 0.345 & 0.224 & 0.505 & 0.877 \\
         EE-DMN~\cite{lin2019entropy} &0.641 & 0.493 & 0.388 & 0.310 & 0.241 & 0.527 & 0.912 \\

         HAN~\cite{sanabria2019cmu} &0.718 & 0.584 & 0.478 & 0.394 & 0.267 & 0.563 & 1.094 \\
         MTN~\cite{le2019multimodal} &0.731 & 0.597 & 0.490 & 0.406 & 0.271 & 0.564 & 1.127 \\
        JMAN~\cite{chu2020multi} &0.667 & 0.521 & 0.413 & 0.334 & 0.239 & 0.533 & 0.941 \\
         MSTN~\cite{lee2020dstc8} & - & - & - & 0.377 & 0.275 & 0.566 & 1.115 \\

         MTN-P~\cite{le2020multimodal} &0.750 & 0.619 & 0.514 & 0.427 & 0.280 & 0.580 & 1.189\\
         BiST~\cite{le2020bist} &0.755 & 0.619 & 0.510 & 0.429 & 0.284 & 0.581 & 1.192 \\
          VGNMN~\cite{le2021vgnmn} & - & - & - & 0.429 & 0.278 & 0.578 & 1.188 \\
         SCGA~\cite{kim2021structured} & 0.745 & 0.622 & 0.517 & 0.430 & 0.285 & 0.578 & 1.201 \\
         \hline
         VGD~\cite{le2020video} &0.750 & 0.621 & 0.516 & 0.433 & 0.283 & 0.581 & 1.196\\
         PDC~\cite{le2021learning} &0.770 & \underline{0.653} & 0.539 & 0.449 & 0.292 & 0.606 & 1.295 \\
         RLM~\cite{li2021bridging}& 0.765 & 0.643 & 0.543 & 0.459 & 0.294 & 0.606 & 1.308\\
         T5RLM~\cite{yoon-etal-2022-information} & 0.767 & 0.644 & 0.542 & \underline{0.461} & 0.296 & 0.608 & 1.311\\
         JVIT~\cite{9897613} & 0.776 & 0.652 & 0.544 & 0.453 & 0.300 & \underline{0.614} & 1.315\\
         DialogMCF~\cite{10147255}  & \underline{0.777} & \underline{0.653} & \underline{0.547} & 0.457 & \underline{0.306} & 0.613 & \textbf{1.352} \\
         \hline
         {ISR} & \textbf{0.792} & \textbf{0.665} & \textbf{0.560} & \textbf{0.472} & \textbf{0.312} & \textbf{0.623} & \underline{1.344}\\
         \hline \hline
    \end{tabular}
    \label{tab:dstc7}
\end{table*}

\section{Experiments}
\subsection{Dataset}
{We employed three publicly available datasets to evaluate our model, including AVSD@DSTC7~\cite{alamri2018audio},  AVSD@DSTC8~\cite{hori2020audio}, and VSTAR~\cite{wang-etal-2023-vstar}. The first two datasets are both derived from the Charades human-activity dataset~\cite{sigurdsson2016hollywood} that captures authentic human interactions about video clips. The training and validation sets in both datasets are the same, while the testing set is completely different. Due to longer sentences in DSTC8, more nuanced utterance comprehension as well as long-term history dependency capture is more critical. Each dialog, tied to a video, contains ten paired question-answer utterances. VSTAR is a large-scale video-grounded dialog understanding dataset, which is collected from 395 TV series (185K 90-second clips) with carefully cleaned dialogs and metadata information. The sheer volume of data exacerbates the complexity of this dataset. Dataset specifics are detailed in Table~\ref{tab:dataset}.}

\subsection{Experimental Settings}

\subsubsection{Implementation Details}
{For model architecture, we fine-tuned the GPT-2 generator and froze the Faster R-CNN model as well as skipping the pre-trained text embedding model.}
We trained our model for 20 epochs using the Adam optimizer. The learning rate was set at 3.5e-4 and reduced tenfold between the 5-th and 10-th epochs. Configurations included a frame count ($F$) of 15, object number ($O$) of 6, and attention heads ($J$) of 3. The feature dimension ($d$) was 768, and batches were organized in sets of 32. During decoding, we employed a beam search with a width of 5, a sequence length capped at 20, and a length penalty of 0.2. {Hyperparameters,
the selection threshold ($p$) and iteration number ($I$), were separately set to 0.6 and 3 using a grid search on the validation set, where the range of $p$ is \{0.0, 0.2, 0.4, 0.6, 0.8, 1.0\} and $I$ ranges from 0 to 5.} Experiments were conducted using PyTorch on a single NVIDIA A100-PCIE-40GB GPU with CUDA 11.0.



\subsubsection{Evaluation Metrics}
We employed widely-accepted automatic evaluation measures for natural language synthesis\footnote{\url{https://goo.su/xzkY}.}. These metrics include BLEU, METEOR, ROUGE-L, and CIDEr. They compare generated responses with ground truth by assessing word alignment and provide insights into fluency, context relevance, and expressive diversity of the produced dialogs.



\begin{table*}[t]
    \renewcommand{\arraystretch}{1.1}
    \caption{Performance comparison with the latest methods on the AVSD@DSTC8 dataset. The best results are highlighted in bold and the second is underlined.}
    \centering
    \begin{tabular}{cccccccc}
        \hline\hline
        Method & BLEU-1 & BLEU-2 & BLEU-3 & BLEU-4 & METEOR & ROUGE-L & CIDEr\\
         \hline
         Baseline~\cite{hori2019end} & - & - & - & 0.293 & 0.212 & 0.483 & 0.679 \\
        MTN~\cite{le2019multimodal} &- & - & - & 0.352 & 0.263 & 0.547 & 0.978 \\
        MDMN~\cite{xie2020audio} & - & - & - & 0.296 & 0.214 & 0.496 & 0.761 \\
        JMAN~\cite{chu2020multi} &0.645 & 0.504 & 0.402 & 0.324 & 0.232 & 0.521 & 0.875 \\


         MSTN~\cite{lee2020dstc8} & - & - & - & 0.385 & 0.270 & 0.564 & 1.073 \\
         MTN-P~\cite{le2020multimodal} & 0.701 & 0.587 & 0.494 & 0.419 & 0.263 & 0.564 & 1.097\\
         STSGR~\cite{geng2021dynamic} & - & - & - & 0.357 & 0.267 & 0.553 & 1.004 \\
         SCGA~\cite{kim2021structured} & 0.711 & 0.593 & 0.497 & 0.416 & 0.276 & 0.566 & 1.123 \\
        \hline
         RLM~\cite{li2021bridging} &0.746 & 0.626 & 0.528 & 0.445 & 0.286 & 0.598 & 1.240\\
        T5RLM~\cite{yoon-etal-2022-information} & 0.749 & 0.631 & 0.529 & 0.445 & 0.290 & 0.600 & 1.263\\
        JVIT~\cite{9897613} & 0.748 & 0.632 & \underline{0.536} & \underline{0.456} & 0.289 & 0.600 & \underline{1.268}\\
        DialogMCF~\cite{10147255}  & \underline{0.756} & \underline{0.633} & 0.532 & 0.449 & \underline{0.293} & \underline{0.601} & 1.253\\
         \hline
         {ISR} & \textbf{0.767} & \textbf{0.645} & \textbf{0.548} & \textbf{0.464} & \textbf{0.304} & \textbf{0.610} & \textbf{1.288}\\
         \hline \hline
    \end{tabular}
    \label{tab:dstc8}
\end{table*}

\begin{table}[!t]
\renewcommand\arraystretch{1.1}
\caption{{Performance comparison with the latest methods on the VSTAR dataset. The best results are highlighted in bold and the second is underlined.} \label{tab:vstar}}
\centering
\begin{tabular}{ccccc}
\hline \hline
Method & BLEU-4 & METEOR & ROUGE-L & CIDEr\\
\hline
OpenViDial coarse~\cite{wang2021modeling} & 0.006& 0.035& 0.063&0.066\\
RLM~\cite{li2021bridging} & 0.010&0.032 & 0.061&0.079 \\
AVDT~\cite{wang-etal-2023-vstar} & \underline{0.014} & \underline{0.040} & \underline{0.080} & \underline{0.137}\\
\hline
ISR & \textbf{0.026}&\textbf{0.051} & \textbf{0.098}&\textbf{0.157} \\
\hline \hline
\end{tabular}
\end{table}

\subsection{Performance Comparison}
{We evaluated the performance of our proposed framework against several state-of-the-art baselines. This comparison includes generalized methods (\textit{e.g.}, MTN~\cite{le2019multimodal}, BiST~\cite{le2020bist}, SCGA~\cite{kim2021structured}), as well as GPT-based ones (\textit{e.g.}, PDC~\cite{le2021learning}, RLM~\cite{li2021bridging}, T5RLM~\cite{yoon-etal-2022-information}) on the AVSD@DSTC7 and AVSD@DSTC8 datasets.} The results for the AVSD@DSTC7 dataset are summarized in Table~\ref{tab:dstc7}, with the following notable insights:
\begin{itemize}
    \item Models leveraging pretrained language constructs like GPT-2 (PDC, RLM, T5RLM, and JVIT) generally outperformed non-pretrained counterparts, {confirming the efficacy of pretrained models in text generation. This also provides a basis for subsequent work using pre-trained model.}
    
    \item Our model surpassed all compared methods, achieving the highest scores across almost all evaluation metrics. 
    {For example, ISR achieves a 4.2\% relative gain in BLEU-4 score compared to the JVIT baseline. These results not only validate the effectiveness of our overall framework but also reveal the necessity of mining deeper clues for question understanding.}
    
\end{itemize}

The performance of our proposed framework, in comparison with several state-of-the-art methods on the AVSD@DSTC8 dataset, is delineated in Table~\ref{tab:dstc8}. Key observations drawn from the comparative results are as follows:
\begin{itemize}
\item  Across most methods, there is a noticeable decrease in performance from AVSD@DSTC7 to AVSD@DSTC8, evidenced by metrics like BLEU-4, METEOR, ROUGE-L, and CIDEr. For example, the BLEU-4 score of RLM drops from 0.459 to 0.445. {
The observed decline in performance may be due to the disparities between the two datasets, particularly because the questions in the AVSD@DSTC8 testing set exhibit greater complexity.
}

\item  {Our proposed model demonstrates significant advancements, achieving improvements across multiple evaluation metrics on the AVSD@DSTC8 benchmark. Notably, it surpasses DialogMCF by 3.3\% in BLEU-4 and by 2.8\% in CIDEr scores. These findings underscore the robustness and enhanced effectiveness of our model.} 
\end{itemize}

{The evaluation results of our ISR model on the VSTAR dataset are shown in Table~\ref{tab:vstar}. And we can derive the following conclusions from this:
\begin{itemize}
    \item The performance of all methods on this dataset is relatively poor. We believe that this is on the one hand caused by the large size of the data, while on the other hand, TV serials are perhaps more difficult for models to understand compared to real videos.
    \item On this extremely difficult dataset, our model also obtains optimal performance compared to the latest baselines. This phenomenon again illustrates the effectiveness of our path search and aggregation strategy and iterative reasoning network for mining the hidden semantics of complex data.
\end{itemize}
}

{In conclusion, our methodology maintains a consistent edge over other contenders on AVSD@DSTC7, AVSD@DSTC8, and VSTAR datasets.} This dominance, mirrored across diverse evaluation metrics, accentuates the dexterity and resilience of our devised approach. 
{In particular, the consistent achievement of top scores across numerous metrics underscores the comprehensive effectiveness of our model for uncovering hidden connections.}
\begin{table*}[t]
    \renewcommand{\arraystretch}{1.1}
    \caption{Performance comparison among the variants of our model on the AVSD@DSTC7 dataset. The best results are highlighted in bold.}
    \centering
    \begin{tabular}{cccccccc}
        \hline\hline
        Method & BLEU-1 & BLEU-4 & METEOR & ROUGE-L & CIDEr\\
         \hline
         {ISR} & \textbf{0.792} & \textbf{0.472} & \textbf{0.315} & \textbf{0.623} & \textbf{1.344}\\
         \hline
          w/o textual encoder& 0.779 & 0.455 & 0.306 & 0.614 & 1.311\\
          w/ single attention & 0.785& 0.467 & 0.311& 0.619& 1.334\\ 
         w/o weight matrix & 0.788&0.469&	0.313&	0.620&	1.335\\

          w/o visual encoder& 0.781 & 0.465 & 0.308 & 0.617 & 1.329\\
          w/o gate mechanism& 0.786 & 0.468 & 0.310 & 0.620 & 1.338\\
          w/o GPT-2 decoder& 0.774 & 0.447 & 0.298 & 0.609 & 1.276\\
         \hline \hline
    \end{tabular}
    \label{tab:ablation}
\end{table*}

\subsection{Ablation Study}
The goal of our ablation study is to critically evaluate the individual contribution of distinct modules within our {ISR} model. By strategically removing certain components, we can assess their significance in the holistic performance of our model. We designed multiple derivatives to accomplish this:
\begin{itemize}
    \item \textbf{w/o textual encoder}. We bypassed the proposed textual encoder, relying solely on the original history features $\mathbf{E}_h$ and question features $\mathbf{E}_{q}^{vis}$ generated by the visual encoder, in lieu of the processed features $\hat{\mathbf{E}}_{h}$ and $\hat{\mathbf{E}}_q$.
    \item \textbf{w/ single attention}. Instead of adopting four separate attention networks, we use only one attention mechanism to obtain $\mathbf{E}_{q}^{q}$, $\mathbf{E}_{q}^{v}$, $\mathbf{E}_{v}^{q}$, and $\mathbf{E}_{v}^{v}$.
    \item \textbf{w/o weight matrix}. Remove the weight matrix obtained in Eqn. (7). This means removing the weight coefficients in Eqn. (9), treating different modalities equally.
    
    \item \textbf{w/o visual encoder}. In this configuration, the proposed visual encoder is omitted, and the initial feature $\mathbf{E}_v$ and question feature $\mathbf{E}_{q}^{txt}$ emanating from the textual encoder are directly inputted into the generator.

    \item \textbf{w/o gate mechanism}. Instead of employing gate fusion, we derived the final question representation $\hat{\mathbf{E}}_q$ through a single fully connected layer, expressed as:
\begin{equation}
    \hat{\mathbf{E}}_q = \mathbf{W}_f[\mathbf{E}_{q}^{txt}||\mathbf{E}_{q}^{vis}]+\mathbf{b},
\end{equation}
where $\mathbf{W}_f \in \mathbb{R}^{{d}\times{2d}}$ and $\mathbf{b}$ symbolize the learnable parameter matrix and vector, respectively.
    \item \textbf{w/o GPT-2 decoder}. In this derivative, the GPT-2 decoder is supplanted with the LSTM-based auto-regressive decoder that is previously utilized in~\cite{le2019multimodal}.
\end{itemize}

Table~\ref{tab:ablation} presents a meticulous evaluation of the {ISR} model against its various derivatives using the AVSD@DSTC7 dataset. 
{We can obtain the following salient insights from the results:}
\begin{itemize}
    \item {The w/o textual encoder variant, which omits the textual encoder, underperforms in comparison to the complete ISR model.} This underscores the criticality of dialog history in grasping the nuances and intent behind user queries.
    
    \item The performance of using a single attention mechanism, \textit{i.e.}, w/ single attention, is significantly worse compared to ISR model with four separate attention networks, thereby validating the rationality and validity of this setup.
    \item The results are significantly lower than those of our ISR method, indicating that treating different modalities equally is inappropriate. This is reasonable, as the contributions of different modalities (\textit{e.g.}, question and video) are not the same.
    \item Omitting the visual encoder, as seen in the w/o visual encoder derivative, diminishes performance, illuminating the pivotal role that iterative reasoning holds in refining question and video representations.
    
    \item {The model that bypasses the gate mechanism (w/o gate mechanism) delivers suboptimal results compared to the ISR.} This result reinforces the effectiveness of the gate mechanism in fusing cross-modal information.
    
    \item There is a marked performance dip in the w/o GPT-2 decoder variant, which is a testament to the instrumental edge that pre-trained language models like GPT-2 provide in handling downstream tasks.
\end{itemize}

Overall, this ablation study demarcates the contribution of each module to the performance of our model, highlighting the robust design of the proposed {ISR} framework.

\begin{table*}[t]
\renewcommand\arraystretch{1.1}
\caption{{Performance comparison with the large video-language models. The version of ChatGPT is ``gpt-3.5-turbo'' and the best performance is highlighted in bold.} \label{tab:vlm}}
\centering
\begin{tabular}{cccccccc}
\hline \hline
\multirow{2}{*}{Methods} & \multicolumn{5}{c}{Objective Metric} &\multicolumn{2}{c}{ChatGPT}\\
 & BLEU-1 &BLEU-4 & METEOR & ROUGE-L & CIDEr & Accuracy & Score\\
\hline
\multicolumn{8}{c}{AVSD@DSTC7}\\
\multicolumn{8}{l}{\textit{Zero-shot Setting}}\\
\hline
Video-LLaMA-7B & 0.360&0.115&0.210 & 0.367&0.383&28.533&2.551\\
Video-LLaVA-7B &0.422&0.142& 0.239 &0.418 &0.447 &33.158&2.918\\
\multicolumn{8}{l}{\textit{Fine-tuning Setting}}\\
\hline
Video-LLaVA-7B &0.749&0.430& 0.287 &0.580 &1.199 &50.318&3.477\\
ISR & \textbf{0.792}&\textbf{0.472}&\textbf{0.315} & \textbf{0.623}&\textbf{1.344}&\textbf{52.783} &\textbf{3.645}\\
\hline
\multicolumn{8}{c}{AVSD@DSTC8}\\
\multicolumn{8}{l}{\textit{Zero-shot Setting}}\\
\hline
Video-LLaMA-7B &0.345&0.110&0.201 & 0.364&0.444&28.737&2.575\\
Video-LLaVA-7B & 0.410&0.137 &0.228 &0.416 &0.527&33.275 &2.998 \\
\multicolumn{8}{l}{\textit{Fine-tuning Setting}}\\
\hline
Video-LLaVA-7B & 0.737&0.410 &0.278 &0.569 &1.142&\textbf{52.336} &\textbf{3.684} \\
ISR & \textbf{0.767}&\textbf{0.464}&\textbf{0.304} & \textbf{0.610}&\textbf{1.288} &51.812&3.658\\
\hline \hline
\end{tabular}
\end{table*}

{
\subsection{Comparison with LVLMs}
It is well known that LVLMs have achieved excellent performance on several video-language tasks, \textit{e.g.}, video question answering and video captioning. Therefore, to verify their effectiveness on the video-grounded dialog task, we chose representative Video-LLaMA~\cite{zhang2023video} and Video-LLaVA~\cite{lin2023video} methods to compare with ours, and the experimental results are shown in Table~\ref{tab:vlm}.}

From the displayed results, under both zero-shot and fine-tuning settings, it is evident that our model significantly outperforms the two LVLMs across all objective metrics on both datasets.
For a broad comparison, we have also utilized ChatGPT for evaluation\footnote{\url{https://github.com/mbzuai-oryx/Video-ChatGPT/tree/main/quantitative_evaluation}.}, focusing on sentence semantics eliminates structure effects. The structured prompt is shown in Fig.~\ref{fig:prompt}.
It should be noted that Accuracy metric refers to whether the predicted answer semantically matches the ground truth, while the Score is an value between 0 and 5, with 5 indicating the highest meaningful match.
In this case, our ISR model similarly beats the fine-tuned Video-LLaVA-7B model on the AVSD@DSTC7 dataset. On the AVSD@DSTC8 dataset, our approach is inferior to the fine-tuned Video-LLaVA-7B model, perhaps because the rich knowledge embedded in LVLM comes into play in the face of more challenging data.

Crucially, compared to 7B parameters of the two LVLMs, our method yields such results using only about 550M parameters (See Table~\ref{tab:table4}).
These findings underscore the challenges involved in adapting LVLMs to video-grounded dialog task and further corroborate the efficacy of our method in uncovering the complex interplays within video-grounded dialog.


\subsection{Effect of Different Encoders/Decoders}

In Table~\ref{tab:3}, we designed several variants to investigate the effects of different encoder (\textit{e.g.}, Faster R-CNN, ResNext, and I3D) and decoder (\textit{e.g.}, GPT-2-Base, LSTM, GPT-2-Large) configurations. From the displayed results, we can observe that modifying the encoders and decoders does not significantly impact the overall model performance, resulting only in minor fluctuations. And in these results, the performance of the variant with Faster R-CNN as encoder and GPT-2-Large as decoder is optimal. Besides, we also found that the variant using lower-quality I3D features could still achieve performance comparable to the optimal baseline (DialogMCF~\cite{10147255}).
These results fully validate the robustness and generalization of our proposed method.

\begin{table*}[h]
    \renewcommand{\arraystretch}{1}
    \caption{Performance comparison among the variants with different encoders and decoders on the AVSD@DSTC7 dataset.}
    \centering
    \begin{tabular}{ccccccc}
        \hline\hline
        Encoder & Decoder & BLEU-1 & BLEU-4 & METEOR & ROUGE-L & CIDEr\\
         \hline
         Faster R-CNN& GPT-2-Base & 0.792 & 0.472 & 0.315 & 0.623 & 1.344\\
         ResNext & GPT-2-Base &0.785	&0.463&	0.310&	0.619	&1.326\\
         I3D & GPT-2-Base &0.780	&0.459	&0.308	&0.616&	1.317\\
         \hline
         Faster R-CNN & LSTM& 0.774 & 0.447 & 0.298 & 0.609 & 1.276\\
         Faster R-CNN & GPT-2-Large&0.805	&0.481&	0.322	&0.634	&1.382\\
         \hline \hline
    \end{tabular}
    \label{tab:3}
\end{table*}

\subsection{Effect of Different Inputs}

In order to explore the impact of the different inputs of generator, we eliminated them individually, as shown in Table~\ref{tab:5}. This will result in three variants (with the question not being removed): w/o video, w/o caption, and w/o history. From the results, we can observe that removing any input leads to a significant decline in model performance. Moreover, the impact of removing different inputs varies, with the influence decreasing in the following order: caption, dialog history, and video. This phenomenon has also been validated in related studies~\cite{10147255, le2019multimodal}.

\begin{table*}[h]
    \renewcommand{\arraystretch}{1}
    \caption{Performance comparison among the variants with different inputs of generator on the AVSD@DSTC7 dataset. The best results are highlighted in bold.}
    \centering
    \begin{tabular}{cccccc}
        \hline\hline
         Variant & BLEU-1 & BLEU-4 & METEOR & ROUGE-L & CIDEr\\
         \hline
         ISR & \textbf{0.792} & \textbf{0.472} & \textbf{0.315} & \textbf{0.623} & \textbf{1.344}\\
         \hline
         w/o video&0.775&0.456&0.307	&0.612	&1.316\\
         w/o caption& 0.757&0.439&0.293&0.603&	1.288\\
         w/o history&0.768 &0.441& 0.302 &0.607	&1.301\\
         \hline \hline
    \end{tabular}
    \label{tab:5}
\end{table*}

\subsection{Decoding Methods}
{Table~\ref{tab:my_label} furnishes an in-depth analysis of various decoding strategies deployed within the domain of video-grounded dialog.} A synopsis of our findings is delineated below:
\begin{itemize}
    \item \textbf{Greedy Search}. 
    While greedy search that selects the most probable subsequent word in each instance exhibits laudable performance (BLEU-1 score of 0.765), {it can sometimes overlook globally coherent or intricate responses due to its short-sighted selection strategy.}

    \item \textbf{Nucleus Sampling}. Posting a BLEU-1 score of 0.702, nucleus sampling amalgamates randomness by selecting from a subset of the most likely subsequent words. {The introduction of diversity in responses might interfere with answer coherence in comparison to its counterparts within video-grounded dialogs.}
    
    \item \textbf{Contrastive Search}. {Achieving a commendable BLEU-1 score of 0.780, the contrastive approach evaluates divergent options, potentially aiding in discerning the nuances of video-grounded responses. Despite its superiority over the first two methods, it could not surpass beam search in our assessment.}
    
    \item \textbf{Beam Search}. {Clinching the pinnacle with a BLEU-1 score of 0.792, beam search meticulously evaluates multiple trajectories during decoding,
    ultimately selecting the most contextually rich sequence. 
    Its dominance across all evaluations attests to its suitability for tasks requiring precise alignment with videos.}

    
\end{itemize}

{A notable feature of video-grounded dialog is its deterministic response modality, firmly anchored in the video and its caption. This deterministic nature differs from the more fluid and open-ended nature of general open-domain dialogs. It's evident, in light of this characteristic, that the beam search mechanism emerges as particularly adept for decoding tasks within this situation.}



\begin{table*}[t]
    \renewcommand{\arraystretch}{1.1}
    \caption{Performance comparison among the variants of different generation methods on the AVSD@DSTC7 dataset. The best results are highlighted in bold.}
    \centering
    \begin{tabular}{cccccccc}
        \hline\hline
        Method & BLEU-1 & BLEU-2 & BLEU-3 & BLEU-4 & METEOR & ROUGE-L & CIDEr\\
         \hline
          Greedy Search & 0.765 & 0.628 & 0.514 & 0.422 & 0.300 & 0.605 &1.254\\
          Nucleus Sampling & 0.702 & 0.544 &0.422 & 0.328 & 0.268 & 0.545 & 0.994\\
          Contrastive Search & 0.780 & 0.649 & 0.545 & 0.447 & 0.307 & 0.606 & 1.312\\
          Beam Search & \textbf{0.792} & \textbf{0.665} & \textbf{0.560} & \textbf{0.472} & \textbf{0.312} & \textbf{0.623} & \textbf{1.344}\\

         \hline \hline
    \end{tabular}
    \label{tab:my_label}
\end{table*}

\begin{figure*}[t]
\centering  
\subfloat[The number of dialog turns.]{
    \includegraphics[width=0.48\linewidth]{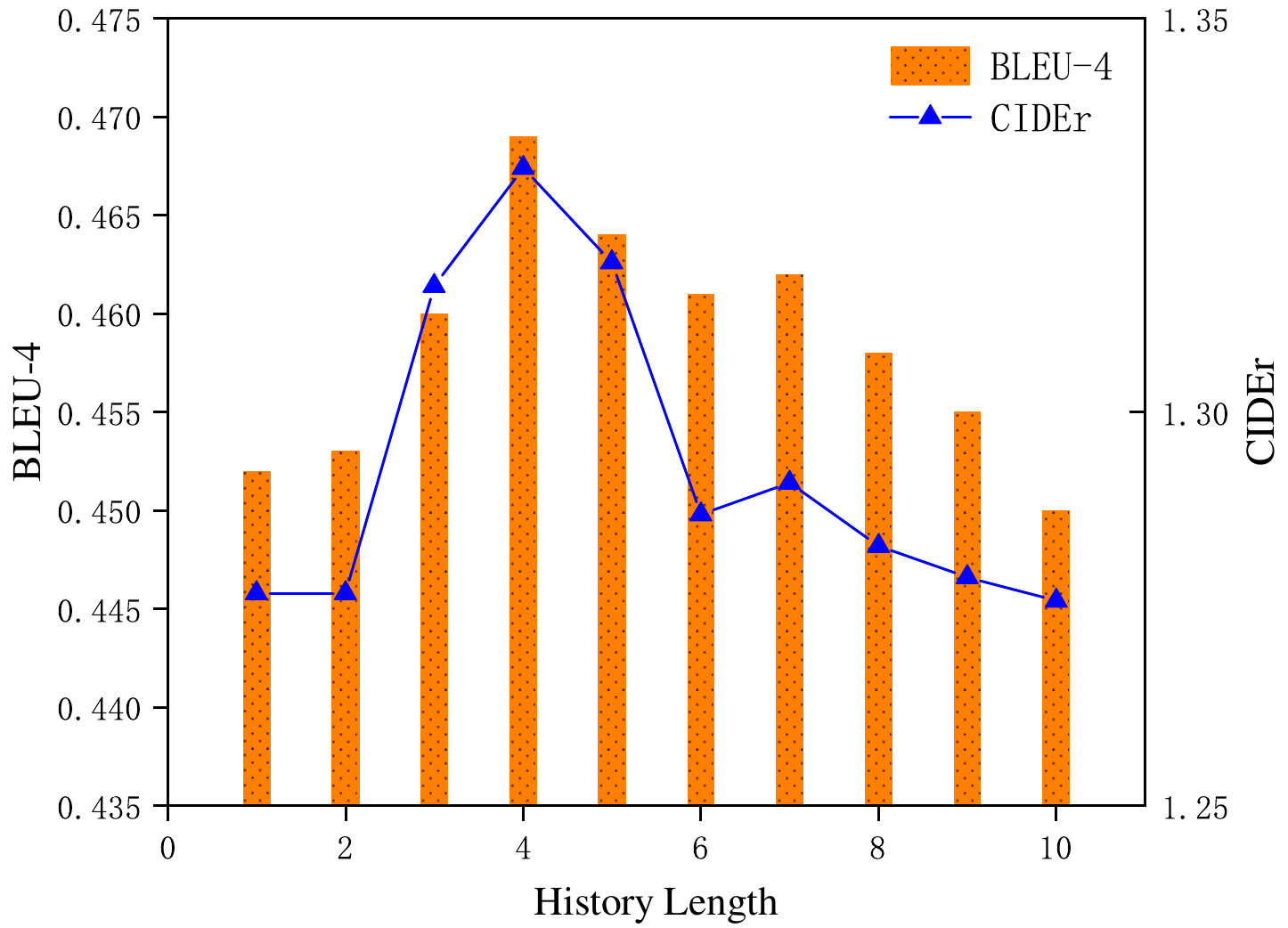}
}
\hfill
\subfloat[The selection threshold $p$.]{
    \includegraphics[width=0.48\linewidth]{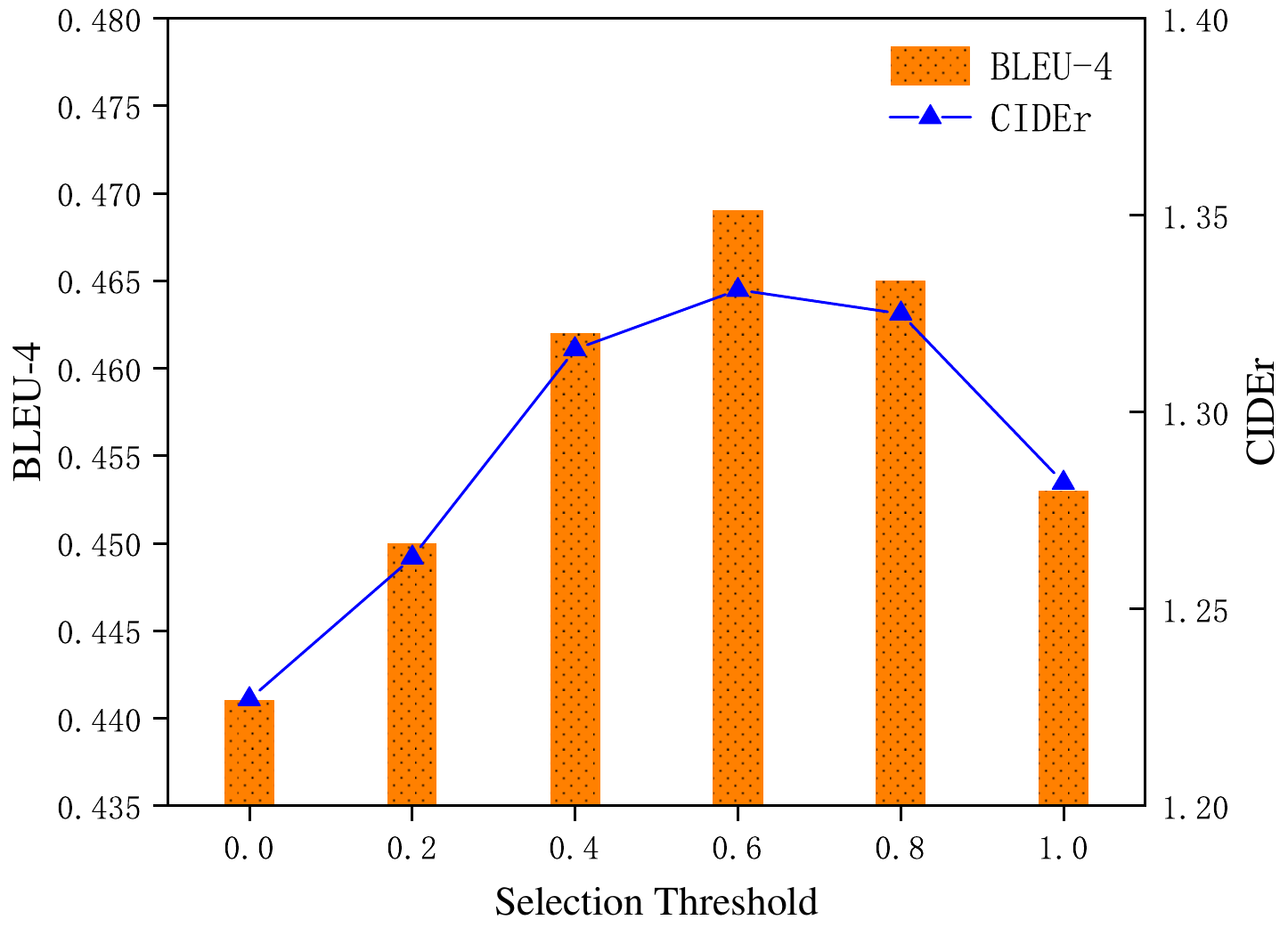}
}
\vspace{1ex}
\subfloat[The number of iterations $I$.]{
    \includegraphics[width=0.48\linewidth]{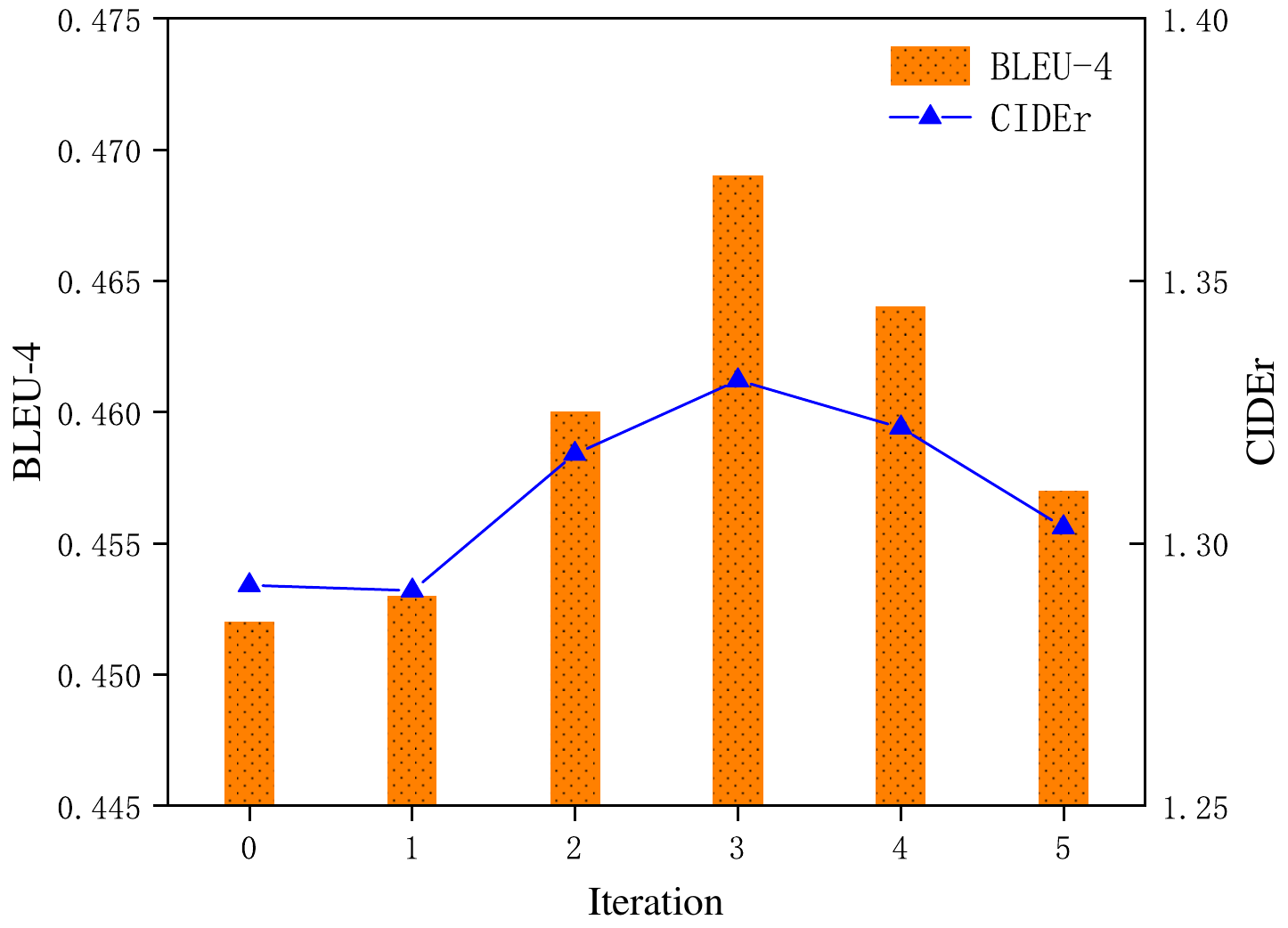}
}
\caption{Parameter analysis of our model in terms of (a) the number of dialog turns, (b) the selection threshold $p$, (c) the number of iterations $I$.}
\label{fig:parameter}
\vspace{-1em}
\end{figure*}



\subsection{Parameter Analysis}
{In this section, we explored in depth the implications of three key hyperparameters: the number of dialog turns, the selection threshold $p$, and the number of iterations $I$ in the ISR model.}
\subsubsection{On the number of dialog turns} 
Dialog turns in our model encapsulate the iterative interaction history, serving as a critical backdrop against which current interactions are contextualized.
{We undertook a quest to ascertain the ideal number of dialog turns for maximizing our model's performance.}

As showcased in Fig.~\ref{fig:parameter}(a), our experiments spanned dialog histories ranging from 1 to 10 turns. A discernible trend emerged: both the BLEU-4 and CIDEr metrics initially surged, peaking at a dialog history length of 4, before exhibiting a downturn. This oscillation underscores a salient feature: while history is undeniably beneficial, inundating the model with an overabundance of it seems counterproductive. The decline in performance with longer dialog histories hints at the introduction of noise or redundant information, which potentially obfuscates more pertinent recent interactions.


\subsubsection{On the selection threshold $p$} 
{The selection threshold $p$ plays a pivotal role in filtering and modulating the information processed by the ISR model.} To pinpoint its optimal value, we scrutinized the performance of our model across varying threshold levels.

{As illustrated in Fig.~\ref{fig:parameter}(b), the CIDEr metric, along with the BLUE-4 score, starts off with a surge, culminating in a peak at a threshold of 0.6, post which there is a decline.} This showcases that extreme values on either end of the spectrum are not conducive. While an extremely stringent threshold might deprive the model of necessary data, an overly lenient one might flood it with superfluous information.


\subsubsection{{On the number of iterations $I$}} 
The reasoning network, nested within the visual encoder, embarks on an iterative journey to refine its understanding of the given content. The question then arises: how many iterations strike the right balance between refinement and over-complication?

Our data, highlighted in Fig.~\ref{fig:parameter}(c), offers an answer. Both BLEU-4 and CIDEr metrics register peak performances at three iterations, suggesting that this is the sweet spot. {A deeper examination reveals that beyond this point, adding more iterations seems counterproductive.} This could be attributed to the additional computational overhead and potential for overfitting that come with more iterations.


\begin{table}[!t]
\renewcommand\arraystretch{1.1}
\caption{Comparison on the number of trainable parameters, memory usage, and inference speed. \label{tab:table4}}
\centering
\begin{tabular}{ccccc}
\hline \hline
Methods & Parameter & Memory & Speed\\
\hline
MTN~\cite{le2019multimodal} &289.11M & 2.59G &3.42s\\
BiST~\cite{le2020bist} &17.99M & 8.08G &1.31s\\
RLM~\cite{li2021bridging} &523.75M& 2.27G& 1.24s \\
\hline
ISR & 546.09M & 2.31G & 1.27s\\
\hline \hline
\end{tabular}
\end{table}

\subsection{Efficiency Analysis}
In order to more fully assess the efficiency of our model, we have calculated the number of trainable parameters, memory usage, and inference speed of our model as well as the representative baselines, which are presented in Table~\ref{tab:table4}. The entire computation is done on AVSD@DSTC7 dataset using a NVIDIA-V100-32G GPU. It is worth noting that since existing work does not emphasize the above factors, we only have compared open-source methods. As reported in Table~\ref{tab:table4}, in comparison to the similar method RLM~\cite{li2021bridging}, our method ISR achieves better performance without significantly increasing memory usage and inference speed.

\begin{figure*}[h!]
  \centering
  \includegraphics[width=0.8\linewidth]{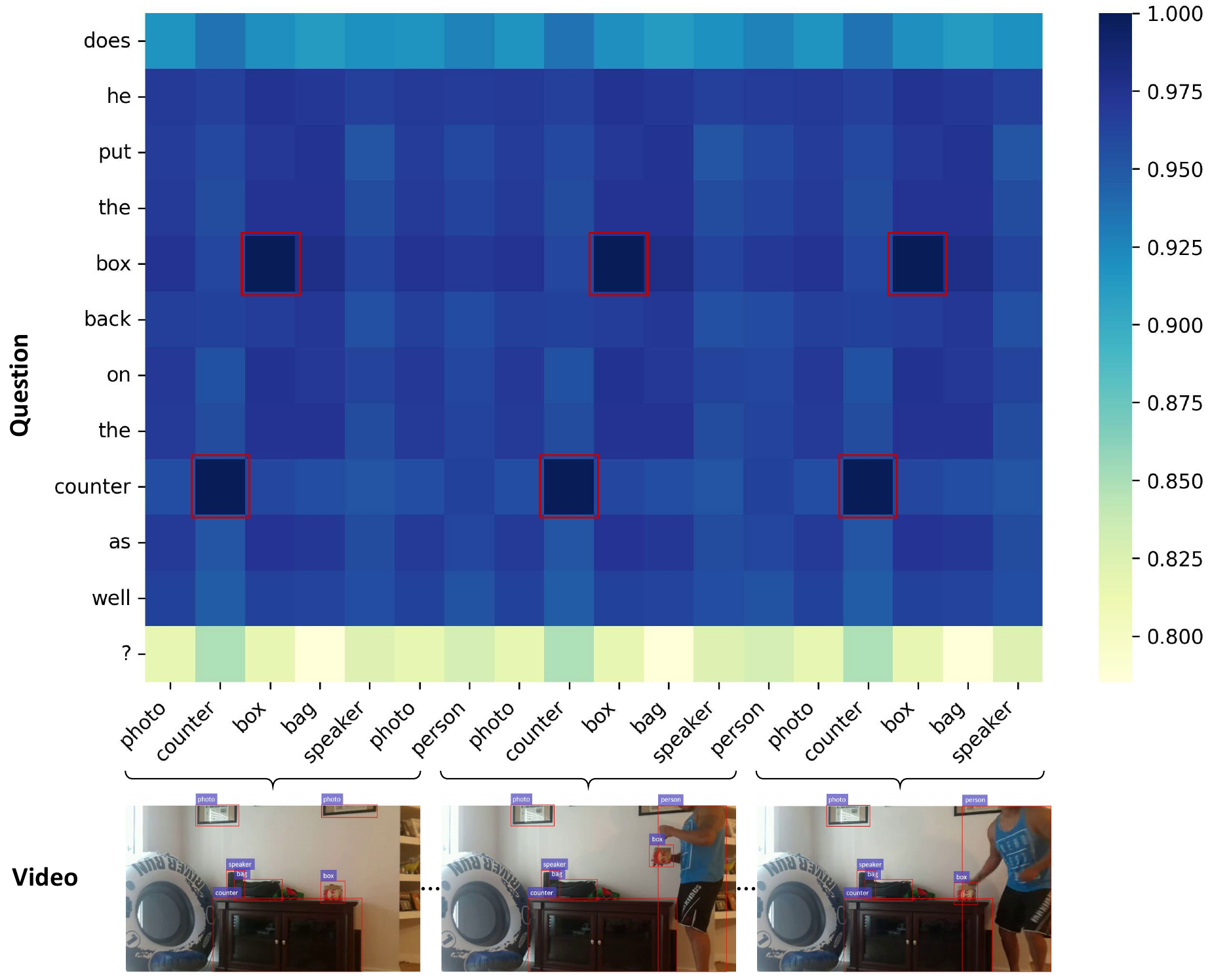}
  \caption{Visualization of question-video weights.}\label{fig:2}
\end{figure*}

\begin{figure*}[t]
\centering  
\subfloat[Case A.]{
    \includegraphics[width=0.6\linewidth]{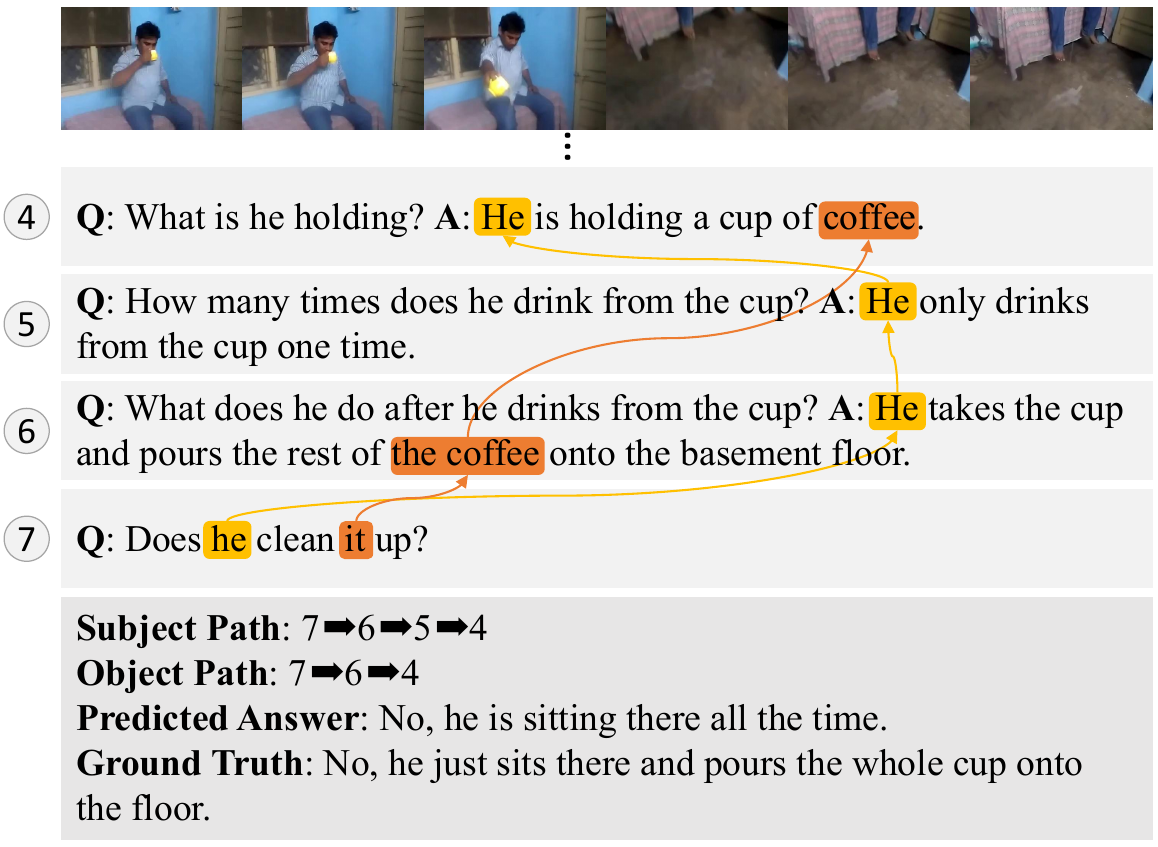} 
}
\hfill
\subfloat[Case B.]{
    \includegraphics[width=0.6\linewidth]{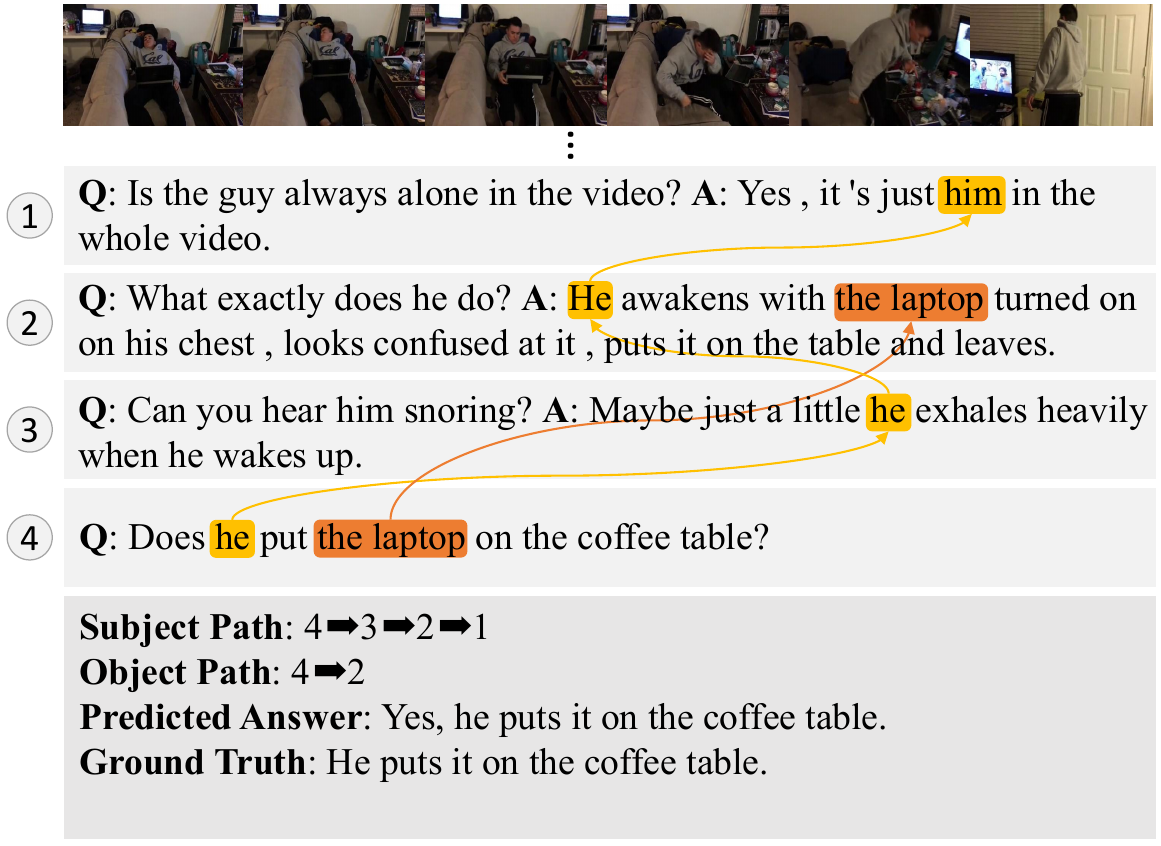} 
}
\caption{Two visualization examples.}   
\label{fig:vis}
\vspace{-1em} 
\end{figure*}


\subsection{Visualization of Question-Video Weights}

To visually present the validity of visual content comprehension, we provided a case and visualized the question-video weights for demonstration, as shown in Figure~\ref{fig:2}. In this case, the question ``does he put the box back on the counter as well ?'' is represented along the Y-axis. For the video, three sampled frames are displayed due to space constraints. Each frame contains the bounding boxes of six objects (fixed setting) along with their corresponding categories, which are mapped to the X-axis. Each grid cell represents the weight between the embedding of a given word in the question and the corresponding object feature in the video frame. From the case, we can observe that word embedding and object feature corresponding to the same concept, such as ``counter'' and ``box'' highlighted in red boxes, exhibit similarity. This provides convincing evidence to demonstrate the validity of the proposed iterative reasoning network.

\subsection{Case Study}
{Delving deeply into the applied implications of our ISR model, we present a series of case studies sourced from the AVSD@DSTC7 dataset.} While quantitative evaluations provide insights into the overall performance of our model, case studies offer a qualitative lens to appreciate its nuanced behaviors and decision-making rationale. {In Fig.~\ref{fig:vis}, we offer a visual presentation of two cases: Case A and Case B.} Each case elucidates the action of our model by illustrating the subject path and object path that the model chooses to traverse en route to deriving the final response.


In Case A, our model demonstrates its competence in sifting through the dialog history, isolating the information salient to the current interaction. It illustrates how the model constructs a subject path that accurately reflects the central theme of the inquiry, as well as an object path that pinpoints the specifics the user is inquiring about. The concordance between these paths and the final response manifests the coherence and logic embedded within the operations of the model. The correct answer is derived by linking the information from the paths, showcasing the intelligent decision-making of our model. {Moreover, the traceable path structure affirms the interpretability of the ISR model.} This fosters a level of trust, as the operations within the model can be tracked and understood rather than being concealed as a ``black box''.

Case B further reinforces the findings from Case A, highlighting the consistency in the performance of our model across diverse scenarios. Here too, the model meticulously constructs paths that are reflective of the dialog history and the user's inquiry. The synthesis of these paths into an accurate answer once again illustrates the ability of our model to harness complex, multi-turn dialogs into coherent, precise responses. The systematic construction and the clear linkage between paths and answers in Case B underscore the robustness of the model and its potential to handle a wide array of real-world queries.




\section{Conclusion and Future Work}
{In this paper, we introduce the Iterative Search and Reasoning (ISR) framework tailored for video-grounded dialog.} The {ISR} framework comprises three main components: a textual encoder, a visual encoder, and a generator. Our textual encoder is equipped with a {path search and aggregation strategy}, which meticulously aggregates context pertinent to the user's question, thus augmenting textual comprehension. In the video domain, we design a multimodal iterative reasoning network within our visual encoder. {This network delves deeply into visual semantics that resonates with the user's inquiry, thereby bolstering visual understanding.} Merging these enriched insights, we employ the pre-trained GPT-2 model as our response generator to produce the final response. {Rigorous evaluations across three datasets attest to the superiority of our ISR model over existing techniques.}


In the future, we intend to explore multimodal large language models, facilitating research in the field of video-grounded dialog. Besides, we also plan to study video-grounded dialog in more perspectives, \textit{e.g.}, egocentric video-grounded dialog.

\begin{acks}
This work is supported in part by the National Natural Science Foundation of China, No.:62376140 and No.:U23A20315; 
Shenzhen College Stability Support Plan, No.:GXWD20220817144428005; the Science and Technology Innovation Program for Distinguished Young Scholars of Shandong Province Higher Education Institutions, No.: 2023KJ128; {Pengcheng Laboratory Major Research Projects, No.:PCL2023A08;} and in part by the Special Fund for Taishan Scholar Project of Shandong Province.
\end{acks}

\bibliographystyle{ACM-Reference-Format}
\bibliography{sample-base}


\end{document}